\documentclass{article}

\usepackage{iclr2027_conference,times}
\iclrfinalcopy

\usepackage{microtype}
\usepackage{graphicx}
\usepackage{subcaption}
\usepackage{booktabs}
\usepackage{multirow}
\usepackage{makecell}
\usepackage{hyperref}
\usepackage{url}
\usepackage{amsmath}
\usepackage{amssymb}
\usepackage{amsfonts}
\usepackage{nicefrac}
\usepackage{xcolor}
\usepackage{tikz}
\usetikzlibrary{positioning}
\usetikzlibrary{decorations.pathreplacing}
\usetikzlibrary{arrows.meta}
\usepackage{pgfplots}
\pgfplotsset{compat=1.18}

\newcommand{\ours}{\textsc{SoftPrompt}}
\newcommand{\lora}{\textsc{LoRA}}
\newcommand{\distill}{\textsc{Soft2Hard}}

\newcommand{\hard}{\textsc{HardPrompt}}
\newcommand{\qwen}{Qwen3-VL}
\newcommand{\rftwenty}{Roboflow20-VL}
\newcommand{\eg}{\textit{e.g.}}
\newcommand{\ie}{\textit{i.e.}}
\definecolor{ours_color}{RGB}{22, 96, 165}
\newcommand{\best}[1]{\textbf{#1}}

\definecolor{findingframe}{RGB}{38,148,132}
\definecolor{findingback}{RGB}{233,247,244}
\newsavebox{\findingbox}
\newenvironment{finding}
  {\par\smallskip\noindent\begin{lrbox}{\findingbox}%
   \begin{minipage}{\dimexpr\linewidth-2\fboxsep-2\fboxrule\relax}\small}
  {\end{minipage}\end{lrbox}%
   \setlength{\fboxrule}{0.9pt}%
   \noindent\fcolorbox{findingframe}{findingback}{\usebox{\findingbox}}\par\smallskip}

\title{ \begin{center} {\LARGE \begin{tabular}{c c c} \multirow{2}{*}{\includegraphics[height=2.2em]{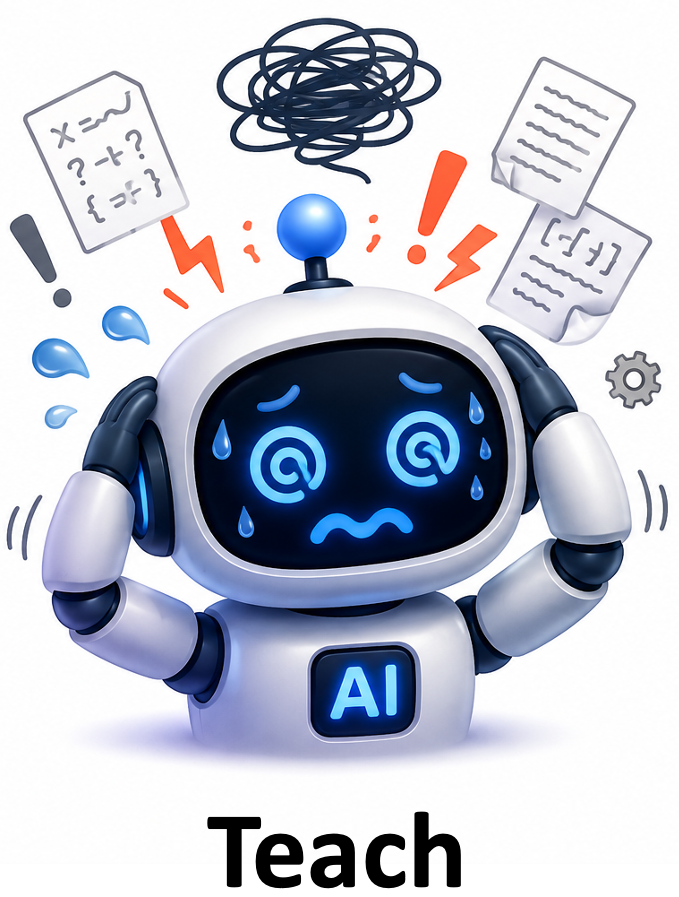}} & Your Model Already Knows & \multirow{2}{*}{\includegraphics[height=2.2em]{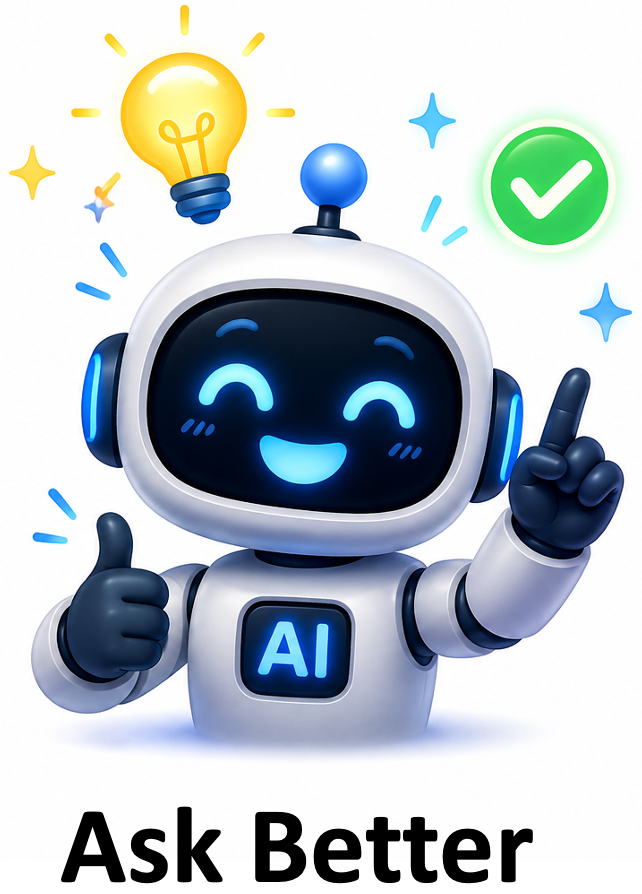}} \\ & Don't Teach It, Learn to Ask It & \end{tabular} } \vspace{0.5em} {\Large Soft Prompting for Few-Shot Adaptation of\\[-1.em] Vision-Language Models} \end{center} } 

\author{ \vspace{0.8em} \parbox{\textwidth}{\centering Gautam Rajendrakumar Gare$^{1}$ \quad Siyi Li$^{1}$ \quad Hewei Wang$^{2}$ \quad Cesar Daniel Hernandez$^{2}$ \\ [0.3em] Wei Zhao$^{2}$ \quad Wolfgang M. Pauli$^{2}$ \quad John Galeotti$^{1}$ \quad Deva Ramanan$^{1}$ \\ [0.8em] $^{1}$Carnegie Mellon University \\ $^{2}$Apple } }

\begin{document}

\maketitle

\begin{abstract}
We address few-shot object detection with vision-language models (VLMs) in out-of-domain settings such as aerial, industrial, and medical imagery, where only ten annotated images are available for supervision.
The dominant adaptation strategies are discrete prompt optimization, which is gradient-free, and \lora{} fine-tuning, which augments frozen pretrained weights with learnt low-rank update matrices.
We revisit a third option: soft prompting, where a few continuous prompt tokens are optimized by gradient descent against a fully frozen backbone.
We show that two design choices strongly influence performance.
First, injecting the tokens at the cross-modal boundary, between the visual and text tokens, beats every other ordering of image, prompt, and text, including the prefix placement inherited from NLP ($10.0$ vs.\ $8.4$ average mAP).
Second, initializing the prompt from a semantically empty space token, so that the prompt's meaning is unchanged at initialisation, beats semantic and random starts.
With these choices, a prompt of one to three learnt tokens, sized per domain by a short sweep ($7{,}168$ parameters on average), matches the best \lora{} configuration from a rank sweep on the \rftwenty{} benchmark ($14.2$ vs.\ $14.2$ mAP, 10-shot) while training over $20{,}000\times$ fewer parameters. However, optimization remains challenging in the few-shot regime: soft prompting exhibits higher variance across random seeds than \lora{}, although \lora{} is itself not immune to instability. In contrast to \lora{}, \ours{} exhibits no forgetting: the \lora{} rank that matches our accuracy loses $35\%$ relative VQA accuracy on NaturalBench, growing to $56\%$ at the largest rank, while \ours{} provably loses none. We attribute this behavior to the fact that \ours{} leaves the pretrained model weights unchanged.
The learnt tokens behave like prompts, not weights: they transfer to a newer model version without retraining ($+0.8$ mAP on Qwen3.5-9B) and can be verbalised into human-readable prompts competitive with dedicated prompt search (matching DetPO, outperforming GEPA).
The recipe extends beyond detection: on RoboCasa manipulation tasks the frozen $\pi_{0.5}$ vision-language-action policy struggles; soft prompts again deliver substantial gains, matching the coverage-matched \lora{} baseline on two of three tasks, provided the tokens are again placed where the gradient needs them; 
Our results suggest that modern VLMs already encode much of what specialised domains require; the task is not to teach them, but to learn how to ask.
\end{abstract}

\section{Introduction}
\label{sec:intro}

\begin{figure}[t]
    \centering
    \includegraphics[width=0.82\textwidth]{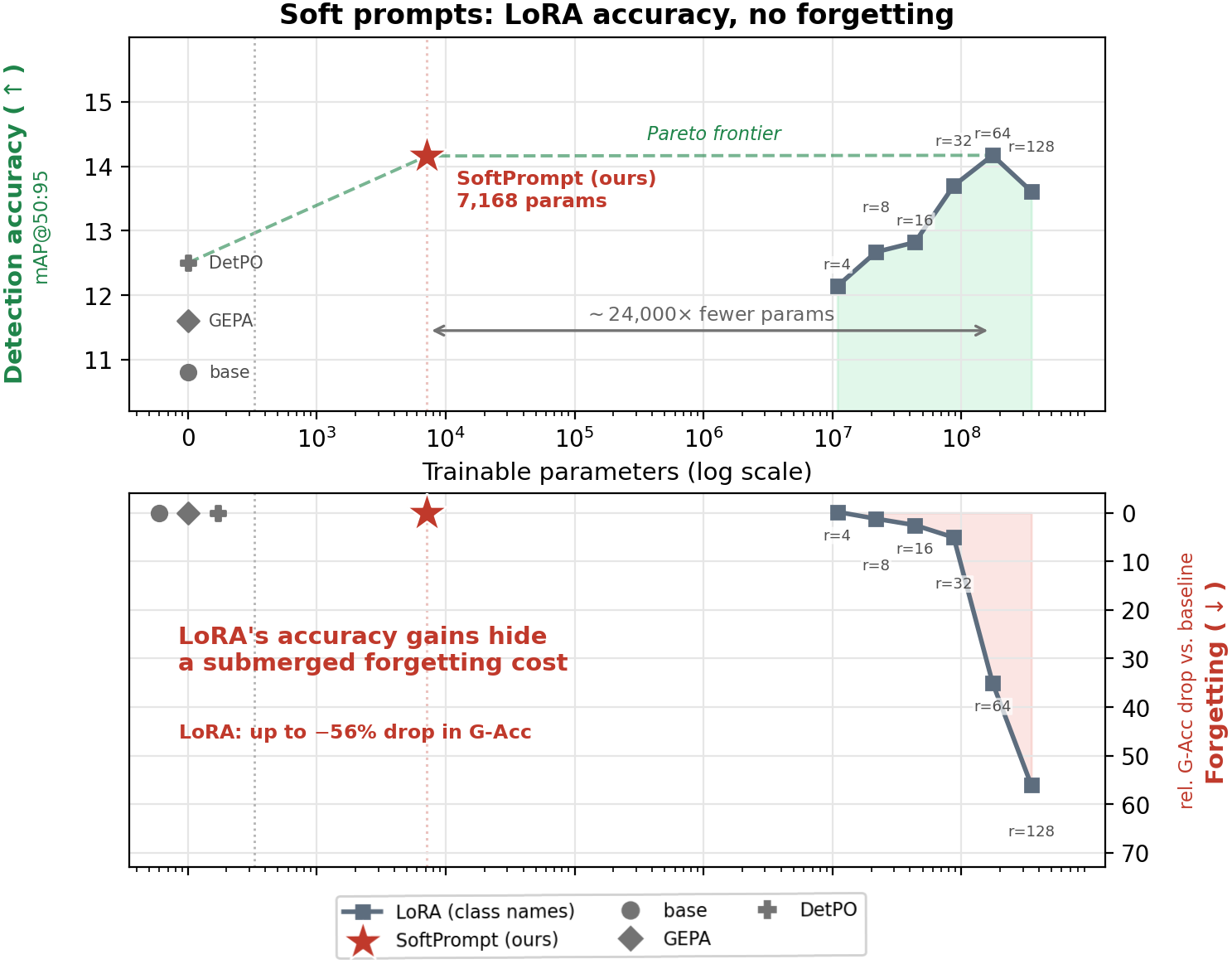}
    \caption{%
    \textbf{\ours{} matches \lora{} with zero forgetting at over $20{,}000\times$ fewer trainable parameters.}
    All methods use the default class-name evaluation protocol (\S\ref{subsec:setup}).
    \emph{Top:} all-domain mAP@50:95 on \rftwenty{} (10-shot) vs.\ trainable parameters (log scale).
    Sweeping \lora{} rank $4\!\to\!128$ peaks at $14.2$ mAP at $r{=}64$ ($174.6$M parameters) and then declines; \ours{} (IST, per-domain $L$, $7{,}168$ parameters on average, backbone frozen) reaches the same $14.2$.
    \emph{Bottom:} the hidden cost.
    Relative drop in NaturalBench VQA group accuracy vs.\ the unadapted backbone: \lora{}'s forgetting grows with rank, $35\%$ at the accuracy-matching $r{=}64$ and $56\%$ at $r{=}128$, while \ours{} and the training-free prompt baselines sit at exactly $0\%$ because no backbone weight changes.
    The two panels move in opposite directions: the ranks that help detection most damage the generalist most.
    The instruction-augmented \lora{} variant is reported in Appendix~\ref{app:lora_sweep} (see \S\ref{subsec:main}).
    Beyond efficiency, the learnt soft prompts keep the virtues of text prompts: they transfer across models (\S\ref{subsec:transfer}) and can be distilled into editable natural language (\S\ref{subsec:distill}).}
    \label{fig:pareto}
\end{figure}

\begin{figure}[t]
    \centering
    \includegraphics[width=0.95\textwidth]{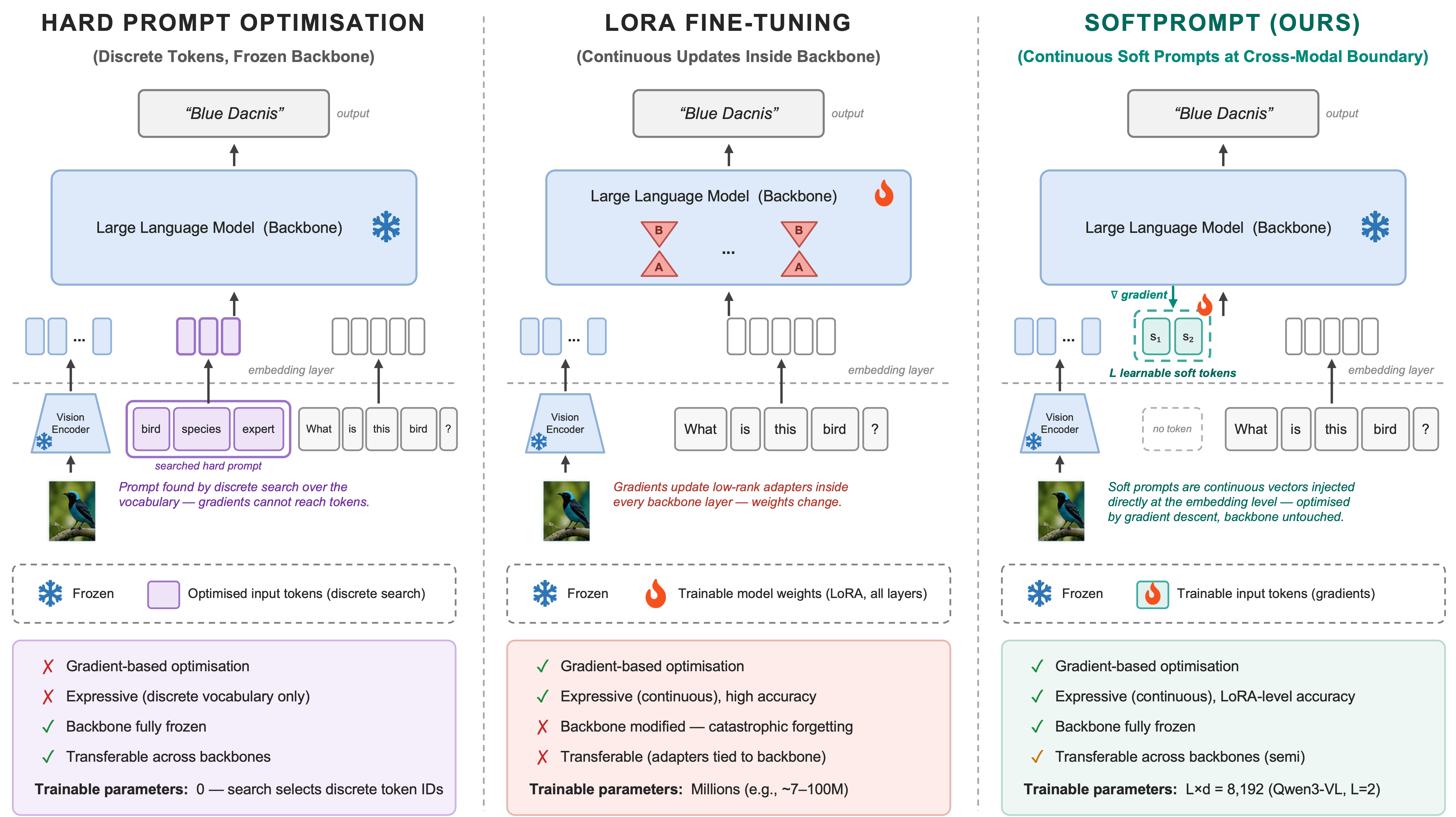}
    \caption{\textbf{\ours{} combines the strengths of the two dominant adaptation approaches.}
    \emph{Hard prompt optimisation} (left) keeps the backbone frozen but searches a discrete vocabulary, which limits expressiveness on specialised domains and precludes gradient-based optimisation.
    \emph{\lora{} fine-tuning} (centre) enables gradient-based adaptation and high accuracy, but modifies millions of backbone weights and risks catastrophic forgetting.
    \emph{\ours{}} (right) optimises $L$ continuous tokens at the cross-modal boundary of the multimodal sequence, reaching \lora{}-level accuracy while keeping the backbone fully frozen and training only $L{\times}d$ parameters ($8{,}192$ at $L{=}2$; $L$ is sized per domain, \S\ref{subsec:length}).
    Like hard prompts and unlike \lora{}, the learnt tokens transfer across models (\S\ref{subsec:transfer}) and can be verbalised into editable text (\S\ref{subsec:distill}).}
    \label{fig:concept}
\end{figure}

Vision-language models (VLMs) trained on web-scale data detect common objects well, but their accuracy collapses on specialised domains: aerial imagery, industrial inspection, medical scans~\citep{Radford2021LearningSupervision,Bai2023Qwen-VL:Beyond,Robicheaux2025Roboflow100-VL:Models}.
In practice, domain experts are expensive and annotation budgets are small, often consisting of 10 labeled images.
The question is how to spend them.

Existing answers fall into two families: Discrete (hard) prompt optimisation and parameter-efficient fine-tuning (PEFT).
Similar to our approach, discrete (hard) prompt optimisation~\citep{AgrawalGEPA:LEARNING,Gare2026DetPO:Detection} also adapts a fully frozen model through prompts rather than weight updates. However, it optimises discrete token sequences using gradient-free search, whereas \ours{} learns continuous prompt embeddings with gradient descent. In our experiments these approaches improve performance at most $1.2$ mAP over a hand-written prompt ($12.5$ vs.\ $11.3$ mAP).
Parameter-efficient fine-tuning, chiefly \lora{}~\citep{Hu2021LoRA:Models}, learns low-rank adaptations to the model parameters. Its rank must be selected per task, and the resulting specialization comes at the expense of general-domain performance.
Figure~\ref{fig:pareto} shows both facts on \rftwenty{}: the \lora{} rank sweep peaks at $14.2$ mAP and then declines, and the same adapters degrade the backbone's VQA accuracy on NaturalBench~\citep{LiNaturalBench:Samples} by up to $56\%$ relative, a cost invisible to detection-only evaluation.

This dichotomy raises a question: \emph{can we have gradient-based adaptation while keeping the pretrained backbone unchanged?}
Soft prompting~\citep{LesterTheTuning} does exactly this: optimising a few continuous token embeddings at the input of a frozen model.
Yet, it is rarely the method of record for multimodal detection.
We find that its weak reputation is largely an artifact of two defaults inherited from NLP: soft tokens are \emph{prepended} to the sequence, and initialised with \emph{task semantics}.
Our experiments show that both defaults are suboptimal in a multimodal sequence.
Injecting the tokens instead at the cross-modal boundary, between the visual patches and the text query, wins our six-way ordering ablation: $10.0$ average mAP vs.\ $8.4$ for the NLP-standard prefix transplant and $6.7$ for the worst ordering, up to $6\times$ on a single dataset (\S\ref{subsec:position}).
Initialising them with the token corresponding to a single space character, so that the prompt's meaning is unchanged at the start, beats every semantic and random initialisation we tried (\S\ref{subsec:init}); the principle is the one behind residual learning~\citep{he2016resnet}, where the safest default for a new module is to change nothing.

With these two choices, the resulting method, which we call \ours{}, is simple: we train $L{=}1$--$3$ learnt tokens, sized per domain ($7{,}168$ parameters on average for an 8B-parameter backbone) with a handful of single-GPU runs.
On \rftwenty{} (10-shot), our result equals the best configuration of a six-rank \lora{} sweep ($14.2$ vs.\ $14.2$ mAP at $r{=}64$), outperforms PEFT across all evaluated ranks, surpasses gradient-free prompt optimisation, and does so with zero forgetting by construction (Figure~\ref{fig:pareto}).
The learnt tokens also keep the practical virtues of text prompts that weight updates lose: they are transferable and interpretable.
If loaded as-is into the newer Qwen3.5-9B, they still improve over that model's zero-shot baseline ($+0.8$ mAP, \S\ref{subsec:transfer}).
They are interpretable in the sense that asking the model to verbalise its own soft tokens yields a human-readable prompt that matches the strongest dedicated prompt search (DetPO) and outperforms GEPA on the same benchmark (\S\ref{subsec:distill}).

Finally, we test whether the recipe is specific to detection.
On three RoboCasa~\citep{nasiriany2024robocasa} manipulation tasks (10 demonstrations each), we adapt a frozen $\pi_{0.5}$ vision-language-action policy~\citep{intelligence2025pi05} with \ours{} (\S\ref{sec:robotics}).
The position lesson reappears in a sharper form: a prefix-only prompt fails ($1.7\%$ success), but a prompt that also reaches the action expert, the module that produces the output, matches or beats coverage-matched \lora{}-based PEFT on the two tasks the base policy can barely do.
The third task, one the base already performs, marks the boundary: the prompt overrides the competent policy while \lora{} improves it.
Where the prompt lives, and whether the frozen model needs to be asked or corrected, decide the outcome.

\textbf{Contributions.}
First, we identify the two levers that make soft prompting competitive for multimodal detection: injection at the cross-modal boundary and a meaning-preserving initialisation, each validated by a dedicated ablation (\S\ref{sec:ablations}).
Second, we compare \ours{} against hard prompts, discrete prompt optimisation, a full \lora{} rank sweep, and a fine-tuned specialist detector on \rftwenty{} under strict 10-shot supervision, and measure what each method costs the generalist backbone on NaturalBench VQA and RefCOCO grounding (\S\ref{sec:experiments}).
Third, we show that the learnt tokens behave like language: we analyse what they encode, distill them into portable hard prompts that beat prompt-search baselines, and quantify their transfer across model versions (\S\ref{sec:analysis}).
Finally, we extend the recipe to a vision-language-action policy across three manipulation tasks and show that the placement principle, not the detection task, is what carries it, and that the frozen model's own competence bounds it (\S\ref{sec:robotics}).

\section{Related Work}
\label{sec:related}

\textbf{VLMs and open-vocabulary detection.}
Modern VLMs such as \qwen{}~\citep{Bai2023Qwen-VL:Beyond}, LLaVA~\citep{Liu2023VisualTuning}, and InstructBLIP~\citep{instructblip} emit detections as structured text.
Specialist open-vocabulary detectors, Grounding DINO~\citep{Liu2023GroundingDetection} and OWL-ViT~\citep{owlvit}, align region features with text embeddings and remain far stronger on raw mAP (Table~\ref{tab:rf20_categories}); they are single-task models and serve as our specialist reference point.

\textbf{Prompt tuning and PEFT.}
Soft prompting originates with prompt tuning~\citep{LesterTheTuning} and prefix tuning~\citep{li2021prefix}, extended to deep layers by P-Tuning~v2~\citep{liu2022ptuningv2}, to vision encoders by VPT~\citep{jia2022vpt}, and to CLIP-style alignment by CoOp/CoCoOp~\citep{ZhouLearningModels,cocoop}, ProDA~\citep{proda}, and MaPLe~\citep{khattak2023maple}.
Like these methods, we train only continuous input tokens, but we are shallow, not deep, prompt tuning: unlike VPT-Deep and P-Tuning~v2, which re-inject a fresh learnable prompt at every transformer layer and so touch the model's internal representations directly, our soft tokens enter the sequence once, at the input, exactly as VPT-Shallow's do, and never again; every other layer's weights and activations remain frozen and untouched by anything trainable.
We also study generative VLM detection, where the prompt sits inside a single autoregressive multimodal sequence and its \emph{position} relative to the two modalities becomes a first-class variable, which we show is decisive (\S\ref{subsec:position}).
\lora{}~\citep{Hu2021LoRA:Models} and adapters~\citep{adapters} are the weight-space alternatives we compare against.
Prompt transfer across models was studied for text-only LMs by SPoT~\citep{vu2022spot}; \S\ref{subsec:transfer} gives the multimodal counterpart.
\citet{alazrakireverse} inject learnable textual preambles into a frozen LLM via reinforcement learning; their preambles live in discrete token space and need a reward signal, whereas our tokens are continuous and trained by plain gradient descent.

\textbf{Hard prompt optimisation.}
GEPA~\citep{AgrawalGEPA:LEARNING} evolves discrete prompt candidates; TextGrad~\citep{yuksekgonul2024textgrad} propagates natural-language critiques as gradient-like optimisation signals, but, unlike \ours{}, does not perform numerical gradient descent; AutoPrompt~\citep{Shin2020AutoPrompt:Prompts} performs gradient-guided discrete search; DetPO~\citep{Gare2026DetPO:Detection} specialises discrete search to detection.
These methods keep the model frozen, as we do, but search a discrete space without gradient access.
We quantify the resulting gap on identical benchmarks (Table~\ref{tab:rf20_categories}) and show that a gradient-trained continuous prompt can be distilled back into their discrete space at a profit (\S\ref{subsec:distill}).

\textbf{Prompt distillation and forgetting.}
PromptKD~\citep{promptkd} distills knowledge between prompts in a teacher-student setup; our \distill{} instead asks the model itself to verbalise its own soft tokens, requiring no student training.
Catastrophic forgetting~\citep{mccloskey1989catastrophic} motivates our cost analysis; our contribution there is measurement (NaturalBench and RefCOCO probes for detection-adapted VLMs), not a new mitigation.

\section{Method}
\label{sec:method}



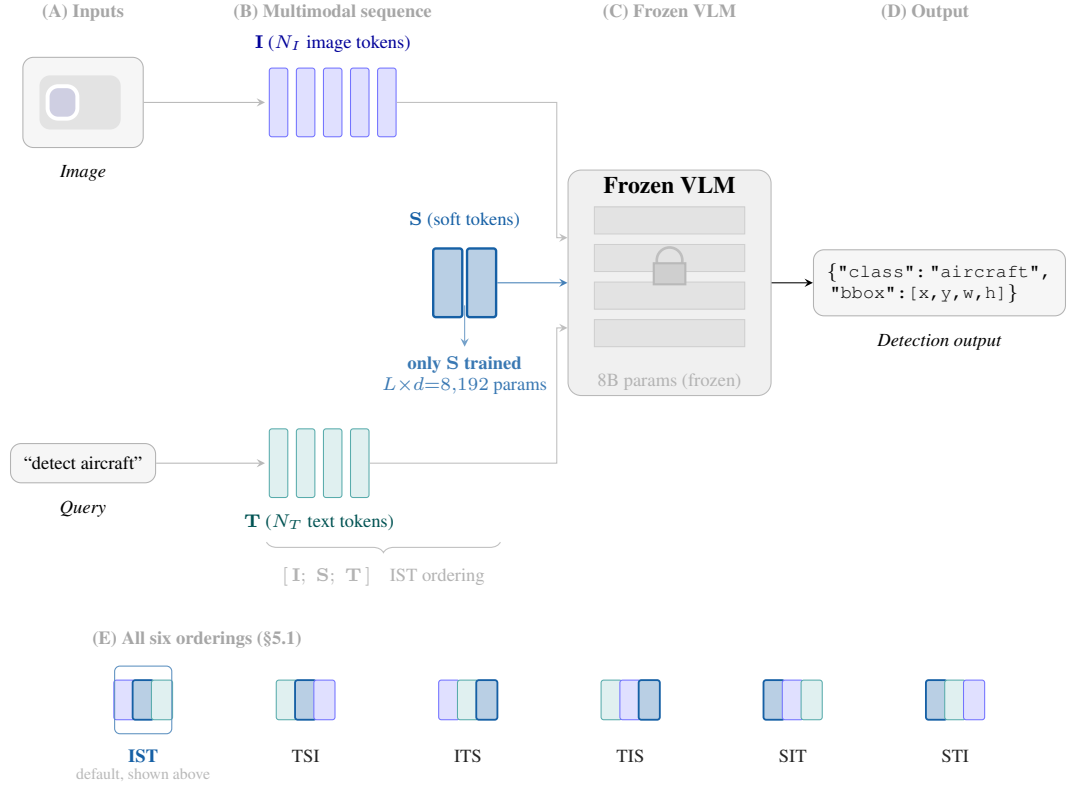
\begin{figure}[t]
  \centering
  \resizebox{\textwidth}{!}{%
  \begin{tikzpicture}[
    font=\small,
    >=stealth,
    vtok/.style = {draw=blue!55, fill=blue!12, rounded corners=1pt,
                   minimum width=7pt, minimum height=26pt, inner sep=0pt},
    ptok/.style = {draw=ours_color, fill=ours_color!30, rounded corners=1pt,
                   minimum width=11pt, minimum height=26pt, inner sep=0pt,
                   line width=0.9pt},
    ttok/.style = {draw=teal!65, fill=teal!12, rounded corners=1pt,
                   minimum width=7pt, minimum height=26pt, inner sep=0pt},
    ibox/.style = {draw=gray!60, fill=gray!7, rounded corners=4pt,
                   inner sep=5pt, align=center},
    vlm/.style  = {draw=gray!55, fill=gray!12, rounded corners=5pt,
                   minimum width=2.7cm, minimum height=3.0cm, align=center},
  ]

  \node[ibox, minimum width=1.6cm, minimum height=1.2cm] (imgbox) at (0, 2.4) {
    \begin{tikzpicture}[scale=0.18]
      \fill[gray!22] (0,0) rectangle (6,4);
      \fill[blue!25!gray!30] (0.6,0.8) rectangle (2.8,3.2);
      \draw[white, line width=1.5pt] (0.6,0.8) rectangle (2.8,3.2);
    \end{tikzpicture}
  };
  \node[font=\scriptsize, below=3pt of imgbox] {\textit{Image}};

  \node[ibox, font=\scriptsize, align=center] (qbox) at (0,-2.4) {``detect aircraft''};
  \node[font=\scriptsize, below=3pt of qbox] {\textit{Query}};

  \node[vtok] (v0) at (2.60, 2.4) {};
  \node[vtok] (v1) at (2.96, 2.4) {};
  \node[vtok] (v2) at (3.32, 2.4) {};
  \node[vtok] (v3) at (3.68, 2.4) {};
  \node[vtok] (v4) at (4.04, 2.4) {};
  \node[font=\scriptsize, blue!60!black, anchor=south] at (3.32, 2.95)
    {$\mathbf{I}$~($N_I$ image tokens)};

  \node[ttok] (t0) at (2.60, -2.4) {};
  \node[ttok] (t1) at (2.96, -2.4) {};
  \node[ttok] (t2) at (3.32, -2.4) {};
  \node[ttok] (t3) at (3.68, -2.4) {};
  \node[font=\scriptsize, teal!65!black, anchor=north] at (3.14, -2.95)
    {$\mathbf{T}$~($N_T$ text tokens)};

  \node[ptok] (p0) at (4.85, 0) {};
  \node[ptok] (p1) at (5.30, 0) {};
  \node[font=\scriptsize\bfseries, ours_color, anchor=south] at (5.075, 0.60)
    {$\mathbf{S}$~\normalfont(soft tokens)};
  \draw[ours_color!60, thin, ->] (5.075,-0.30) -- (5.075,-0.85)
    node[below, anchor=north, align=center, font=\scriptsize, ours_color!85]
    {\textbf{only $\mathbf{S}$ trained}\\$L{\times}d{=}8{,}192$ params};

  \draw[->, gray!55, thin] (imgbox.east) -- ++(0.55,0) |- (2.42, 2.4);
  \draw[->, gray!55, thin] (qbox.east)   -- ++(0.55,0) |- (2.42,-2.4);

  \draw[gray!50, thin, decorate,
        decoration={brace, mirror, amplitude=4pt}]
    (2.44,-3.45) -- (5.55,-3.45)
    node[midway, below=6pt, font=\scriptsize, gray!60]
    {$[\,\mathbf{I};\;\mathbf{S};\;\mathbf{T}\,]$\quad IST ordering};

  \node[vlm] (vlm) at (7.80, 0) {};
  \foreach \k in {0,1,2,3} {
    \fill[gray!22] (6.80,-0.85+\k*0.50) rectangle (8.80,-0.85+\k*0.50+0.36);
    \draw[gray!38, very thin] (6.80,-0.85+\k*0.50) rectangle (8.80,-0.85+\k*0.50+0.36);
  }
  \node[font=\footnotesize\bfseries] at (7.80, 1.30) {Frozen VLM};
  \node[font=\scriptsize, gray!60, align=center] at (7.80,-1.32) {8B params (frozen)};
  \draw[gray!50, line width=1pt] (7.80, 0.30) circle (0.17cm);
  \fill[gray!38] (7.59,-0.02) rectangle (8.01,0.26);
  \draw[gray!55, line width=0.8pt] (7.59,-0.02) rectangle (8.01,0.26);

  \draw[->, gray!50, thin]
    (v4.east) -- (6.30, 2.4) -- (6.30, 0.60) -- (vlm.west |- 0,0.60);
  \draw[->, ours_color!80]
    (p1.east) -- (vlm.west);
  \draw[->, gray!50, thin]
    (t3.east) -- (6.30,-2.4) -- (6.30,-0.60) -- (vlm.west |- 0,-0.60);

  \node[ibox, font=\scriptsize\ttfamily, align=left,
        text width=3.0cm, right=0.55cm of vlm] (outbox)
    {\{"class":\,"aircraft",\\\,"bbox":[x,y,w,h]\}};
  \node[font=\scriptsize, below=3pt of outbox] {\textit{Detection output}};
  \draw[->] (vlm.east) -- (outbox.west);

  \node[font=\scriptsize\bfseries, gray!70] at (   0, 3.60) {(A)~Inputs};
  \node[font=\scriptsize\bfseries, gray!70] at (3.32, 3.60) {(B)~Multimodal sequence};
  \node[font=\scriptsize\bfseries, gray!70] at (7.80, 3.60) {(C)~Frozen VLM};
  \node[font=\scriptsize\bfseries, gray!70] at (11.20,3.60) {(D)~Output};

  \node[font=\scriptsize\bfseries, gray!70, anchor=west] at (0, -4.75)
    {(E)~All six orderings (\S\ref{subsec:position})};

  \node[vtok, minimum width=8pt, minimum height=15pt] at (0.55,-5.55) {};
  \node[ptok, minimum width=8pt, minimum height=15pt, line width=0.7pt] at (0.80,-5.55) {};
  \node[ttok, minimum width=8pt, minimum height=15pt] at (1.05,-5.55) {};
  \draw[ours_color!70, rounded corners=2pt] (0.42,-6.00) rectangle (1.18,-5.10);
  \node[font=\scriptsize\bfseries, ours_color, anchor=north] at (0.80,-6.08) {IST};
  \node[font=\tiny, gray!55, anchor=north] at (0.80,-6.34) {default, shown above};

  \node[ttok, minimum width=8pt, minimum height=15pt] at (2.71,-5.55) {};
  \node[ptok, minimum width=8pt, minimum height=15pt, line width=0.7pt] at (2.96,-5.55) {};
  \node[vtok, minimum width=8pt, minimum height=15pt] at (3.21,-5.55) {};
  \node[font=\scriptsize, gray!30!black, anchor=north] at (2.96,-6.08) {TSI};

  \node[vtok, minimum width=8pt, minimum height=15pt] at (4.87,-5.55) {};
  \node[ttok, minimum width=8pt, minimum height=15pt] at (5.12,-5.55) {};
  \node[ptok, minimum width=8pt, minimum height=15pt, line width=0.7pt] at (5.37,-5.55) {};
  \node[font=\scriptsize, gray!30!black, anchor=north] at (5.12,-6.08) {ITS};

  \node[ttok, minimum width=8pt, minimum height=15pt] at (7.03,-5.55) {};
  \node[vtok, minimum width=8pt, minimum height=15pt] at (7.28,-5.55) {};
  \node[ptok, minimum width=8pt, minimum height=15pt, line width=0.7pt] at (7.53,-5.55) {};
  \node[font=\scriptsize, gray!30!black, anchor=north] at (7.28,-6.08) {TIS};

  \node[ptok, minimum width=8pt, minimum height=15pt, line width=0.7pt] at (9.19,-5.55) {};
  \node[vtok, minimum width=8pt, minimum height=15pt] at (9.44,-5.55) {};
  \node[ttok, minimum width=8pt, minimum height=15pt] at (9.69,-5.55) {};
  \node[font=\scriptsize, gray!30!black, anchor=north] at (9.44,-6.08) {SIT};

  \node[ptok, minimum width=8pt, minimum height=15pt, line width=0.7pt] at (11.35,-5.55) {};
  \node[ttok, minimum width=8pt, minimum height=15pt] at (11.60,-5.55) {};
  \node[vtok, minimum width=8pt, minimum height=15pt] at (11.85,-5.55) {};
  \node[font=\scriptsize, gray!30!black, anchor=north] at (11.60,-6.08) {STI};

  \end{tikzpicture}}
  \caption{%
    \textbf{Overview of \ours{}: $L$ continuous tokens at the cross-modal boundary of a frozen VLM.}
    \textbf{(A)}~An image and a text query are encoded into patch and token embeddings.
    \textbf{(B)}~$L$ learnable soft tokens~$\mathbf{S}$ (blue) are inserted between image tokens~$\mathbf{I}$ (purple) and text tokens~$\mathbf{T}$ (teal), forming $[\mathbf{I};\mathbf{S};\mathbf{T}]$, the IST ordering.
    \textbf{(C)}~The frozen backbone processes the sequence; only $\mathbf{S}$ ($L{\times}d = 8{,}192$ parameters for \qwen{}-8B at $L{=}2$) is trained on the 10-shot support set.
    \textbf{(D)}~Detections are generated as structured text.
    \textbf{(E)}~Simplified token layout for all six orderings ablated in \S\ref{subsec:position} (Table~\ref{tab:placement}): the two cross-modal boundaries (IST, TSI), the two soft-last suffixes (ITS, TIS), and the two soft-first prefixes (SIT, STI); IST (outlined) is the default used in panel~(B).
    Unlike prompt-tuning variants that fine-tune the embedding layer, the embedding matrix itself is untouched: the model gains $L$ new rows and loses nothing.}
  \label{fig:overview}
\end{figure}

Figure~\ref{fig:concept} positions \ours{} between the two existing adaptation families; Figure~\ref{fig:overview} below details the full pipeline and, in panel~(E), every injection ordering ablated in \S\ref{subsec:position}.
The method has three ingredients: where the tokens go, how they are trained, and how they are initialised.
Each is simple; each is validated by an ablation in \S\ref{sec:ablations}.

\subsection{Soft tokens at the cross-modal boundary}
\label{subsec:method_position}

A VLM $f_\theta$ encodes an image into $N_I$ patch embeddings $\mathbf{I} \in \mathbb{R}^{N_I \times d}$ and a text query into $N_T$ token embeddings $\mathbf{T} \in \mathbb{R}^{N_T \times d}$, then decodes bounding boxes autoregressively as structured text.
We introduce $L$ learnable soft tokens $\mathbf{S} \in \mathbb{R}^{L \times d}$ and must choose where in the sequence they go.
The six orderings of image (I), soft tokens (S), and text (T) define the candidate positions: the soft-first prefixes (SIT, the NLP-standard transplant, and STI), the soft-last suffixes (ITS and TIS), and the two cross-modal boundaries (IST and TSI).
We place them between the modalities:
\begin{equation}
  \tilde{\mathbf{X}} = [\mathbf{I};\; \mathbf{S};\; \mathbf{T}]
    \in \mathbb{R}^{(N_I + L + N_T) \times d},
  \label{eq:ist}
\end{equation}
the IST ordering; each ordering acronym is literally the sequence order of $\mathbf{I}$, $\mathbf{S}$, and $\mathbf{T}$.
The intuition is that tokens at this position attend to every visual patch and are attended by every text token, making them a natural bottleneck for a domain-specific visual prior; a prefix token, by contrast, precedes all context and has nothing to condition on.
\S\ref{subsec:position} confirms this intuition: IST beats every prefix and suffix ordering, and neither prefix ever wins an above-floor dataset (one of the ablation pool's eight where scores are not pinned near zero regardless of configuration; the ninth, dentalai, is a floor control, \S\ref{subsec:setup}).
The embedding matrix itself is never modified; it only gains $L$ new rows and loses nothing.

\subsection{Training and what the gradient reaches}
\label{subsec:method_training}

Training minimises the backbone's standard next-token detection loss over the 10-shot support set $\mathcal{D}$, with $\theta$ frozen:
\begin{equation}
  \min_{\mathbf{S}} \; \mathcal{L}_\text{det}(\mathbf{S};\; \mathcal{D}, f_\theta).
  \label{eq:loss}
\end{equation}
The length $L$ is selected per domain by the short sweep of \S\ref{subsec:length}; on \rftwenty{} the selected values are $L \in \{1,2,3\}$ with mean $1.75$ (per-dataset configurations in Appendix~\ref{app:best_config}).
For \qwen{}-8B ($d{=}4096$) this is $4{,}096$--$12{,}288$ trainable parameters, $7{,}168$ on average, roughly $1{,}500\times$ fewer than the smallest \lora{} we evaluate ($r{=}4$) and $24{,}000\times$ fewer than the rank that matches its accuracy ($r{=}64$).

This frugality comes with a distinct optimisation signature, detailed in Appendix~\ref{app:gradflow} (Figure~\ref{fig:gradflow}).
The backward pass is the same computation for both methods: the gradient flows from the loss back through every frozen block.
The difference is where the trainable sinks sit.
\lora{} places one at every depth, so the nearest adapter is one block from the loss and an update lands locally at every layer along the chain; \ours{}'s single sink is the input-side soft tokens, at the far end of the longest backward path, with no update anywhere before it.
The soft tokens do sit in the stream that produces the loss, so the gradient survives and prompt tuning is viable; but this single far-end lever is exactly why \emph{where} the tokens are injected matters so much (\S\ref{subsec:position}), a sensitivity a per-layer weight edit does not share.
The same principle returns, in a sharper form, when we prompt an action policy in \S\ref{sec:robotics}.

\subsection{Initialisation: change nothing at the start}
\label{subsec:method_init}

The third lever is where optimisation starts.
Residual learning~\citep{he2016resnet} made deep networks trainable by making the default behaviour of every layer the identity; we apply the same principle to the prompt.
We initialise all soft tokens with the embedding of a single space character, a token that leaves the prompt's meaning unchanged.
The model's behaviour therefore starts exactly at the (already sensible) frozen baseline, and optimisation moves away only where the loss demands it.
A semantic initialisation, such as class names or a task description, instead \emph{starts} by changing what the prompt says, and the optimiser must first undo any damage before it can help.
\S\ref{subsec:init} shows the space token wins 5 of 8 above-floor pool datasets and beats the frozen baseline on all 8, while the semantically richest initialisations are the least reliable.

\subsection{Soft-to-Hard distillation (\distill{})}
\label{subsec:method_distill}

The soft tokens live in the backbone's own representation space, so the model itself should be able to say what they mean.
\distill{} converts a trained soft prompt into a reusable hard (discrete) prompt in three stages, all served by a single extended-vocabulary engine (the base model plus the trained soft-embedding rows):
\begin{enumerate}
  \setlength\itemsep{0pt}
  \item \textbf{Generate.} For every training image, query the model in its training format and ask it to \emph{describe the objects being detected}; keep the natural-language responses.
  \item \textbf{Summarise.} A text-only pass condenses the per-image descriptions into one global detection instruction.
  \item \textbf{Evaluate.} Insert the distilled instruction into the standard detection template and score the \emph{base} model, with no soft tokens anywhere, on the test split.
\end{enumerate}
To isolate the soft prompt's contribution, we run the identical recipe with descriptions generated \emph{without} the soft tokens (the base source); the soft$-$base gap is what the soft prompt adds (\S\ref{subsec:distill}).

\section{Experiments}
\label{sec:experiments}

\subsection{Setup}
\label{subsec:setup}

\textbf{Benchmark.}
We evaluate on \rftwenty{}, a 20-domain subset of Roboflow-100-VL~\citep{Robicheaux2025Roboflow100-VL:Models} covering all seven of its super-categories (aerial, documents, flora and fauna, industrial, medical, sports, other).
Each domain provides a 10-shot support set; evaluation uses the benchmark's full held-out test split.
We additionally use the LVIS Rare 50 protocol of DetPO~\citep{Gare2026DetPO:Detection} as a long-tail cross-benchmark check (\S\ref{subsec:lvis}).

\textbf{Models.}
All main detection results use frozen \qwen{}-8B-Instruct~\citep{Bai2023Qwen-VL:Beyond}.
Qwen3.5-9B appears only as the target of the cross-model transfer study (\S\ref{subsec:transfer}), Gemma-4-12B-it~\citep{gemma} as an unrelated second backbone trained from scratch (\S\ref{subsec:gemma4}), and $\pi_{0.5}$~\citep{intelligence2025pi05} in the robotics study (\S\ref{sec:robotics}).

\textbf{Baselines.}
\hard{}: the frozen backbone queried with class names, and, as a training-free reference, with hand-written per-class instructions.
GEPA~\citep{AgrawalGEPA:LEARNING} and DetPO~\citep{Gare2026DetPO:Detection}: discrete prompt optimisation on the same frozen backbone.
\lora{}~\citep{Hu2021LoRA:Models}: ranks $4$--$128$; Table~\ref{tab:rf20_categories} shows the best rank and Appendix~\ref{app:lora_sweep} the full sweep, including the instruction-augmented variant discussed in \S\ref{subsec:main}.
Grounding DINO~\citep{Liu2023GroundingDetection}: zero-shot and 10-shot fine-tuned specialist reference.
All generalist methods share the same backbone and the same multi-class evaluation protocol (all classes queried in one call).
Class-name queries are the default protocol throughout; appending hand-written per-class instructions is the variant reported for \hard{} (Table~\ref{tab:rf20_categories}) and \lora{} (Appendix~\ref{app:lora_sweep}), and GEPA and DetPO supply their optimised prompt in place of a hand-written one, which is the method itself.
DetPO's single-class protocol (one class per call, more calls per image) is not cost-comparable and is reported in Appendix~\ref{app:single_class}.

\textbf{Training.}
\ours{}: AdamW, lr $5{\times}10^{-3}$, cosine schedule, batch size 4, up to 5 epochs; space-token initialisation; metrics are COCO-style mAP@50:95 on a 0--100 scale.
Full implementation details, including the \lora{} recipe, are in Appendix~\ref{app:impl}.

\textbf{Ablation pool.}
All ablations (\S\ref{sec:ablations}, Appendix~\ref{app:substitution}) run on a fixed pool of 9 of the 20 datasets, chosen so that every RF20 super-category is represented: aerial-airport (Aerial), actions and lacrosse (Sports), all-elements (Documents), defect-detection (Industrial), dentalai (Medical), flir-camera-objects and new-defects-in-wood (Other), and wb-prova (Flora and Fauna).
The same nine datasets appear in every ablation table and figure.
dentalai, where every method scores ${\sim}0$ (the medical failure discussed in \S\ref{subsec:main}), is kept as the pool's known floor control: no manipulation in any ablation moves it, so it is excluded from ablation win-counts and averages but reported in every table.
Each table carries the frozen-baseline row (class-name query, no soft tokens), taken from the same evaluation that produces Table~\ref{tab:rf20_categories}; the identical values serve as the $L{=}0$ points of the length sweep.

\textbf{Disclosures.}
(i)~Numbers are single runs; we observe a $20$--$28\%$ spread across seeds in this few-shot regime (measured directly: two additional seeds per pool dataset, Appendix~\ref{app:seed_variance}), so per-dataset deltas below ${\sim}1$ mAP should be read as ties, and the headline claim ($14.2$ vs.\ $14.2$) as parity.
(ii)~All main-table \ours{} numbers use the fixed IST position.
A per-dataset best-of-IST/TSI selection reaches $14.7$ but selects the position on the test split; it appears only as the distillation ceiling and transfer origin (\S\ref{sec:analysis}), flagged there, and in Appendix~\ref{app:lora_sweep}.

\subsection{Main results on \rftwenty{}}
\label{subsec:main}

\begin{table}[t]
  \centering
  \caption{\textbf{Two learnt tokens match the best \lora{} fine-tune on \rftwenty{} 10-shot (mAP@50:95).}
    \ours{} reaches $14.2$ all-domain mAP while training $24{,}000\times$ fewer parameters than \lora{} $r{=}64$, beats every training-free prompt baseline by $1.7$--$3.4$ mAP, and wins the Sports and Other categories outright among generalist methods; only the fine-tuned specialist (bottom block) remains ahead, at $2.5\times$ the mAP.
    Domains aggregated into the seven Roboflow-100-VL super-categories; the last column averages all 20 datasets.
    \textbf{Params} is the trainable footprint.
    The \lora{} row shows the best rank of the class-names sweep; the full $r{=}4$--$128$ sweep and the instruction-augmented variant (see \S\ref{subsec:main}) are in Appendix~\ref{app:lora_sweep}.
    \ours{}'s prompt length is chosen per domain ($L \in \{1,2,3\}$, mean $1.75$, hence the $7.2$K average parameter count; per-dataset configurations in Appendix~\ref{app:best_config}).
    \best{Bold}: best per column within the generalist comparison.
    DetPO multi-class numbers are its re-evaluation under the shared evaluation template used by all rows; single-class DetPO variants (not cost-comparable) are in Appendix~\ref{app:single_class}.
    The specialist block bounds the claim: a fine-tuned single-task detector remains far stronger on raw mAP (see the discussion in \S\ref{subsec:main}).}
  \label{tab:rf20_categories}
  \small
  \setlength{\tabcolsep}{4pt}
  \resizebox{\textwidth}{!}{%
  \begin{tabular}{lc|ccccccc|c}
    \toprule
    \textbf{Method} & \textbf{Params} & \textbf{Aerial} & \textbf{Doc.}
                    & \textbf{F.\,\&\,F.} & \textbf{Ind.}
                    & \textbf{Med.} & \textbf{Sports}
                    & \textbf{Other} & \textbf{All} \\
    \midrule
    \multicolumn{10}{l}{\textit{Frozen backbone, prompt-space (training-free)}}\\
    \hard{} (class names)                     & 0 &  6.4 &  7.5 & 25.0 &  8.2 &  0.0 &  8.6 &  9.1 & 10.8 \\
    \hard{} + instructions                    & 0 &  7.4 &  8.5 & 25.0 &  9.1 &  0.2 &  9.7 &  9.6 & 11.3 \\
    GEPA~\citep{AgrawalGEPA:LEARNING}         & 0 &  6.3 & 10.8 & 22.4 &  8.5 &  0.1 & 12.8 & 11.5 & 11.6 \\
    DetPO~\citep{Gare2026DetPO:Detection}     & 0 &  8.6 & 10.1 & 26.7 & \best{9.9} &  0.1 & 11.0 & 10.5 & 12.5 \\
    \midrule
    \multicolumn{10}{l}{\textit{Frozen backbone, prompt-space (gradient-trained, ours)}}\\
    \ours{} (IST, per-domain $L$)             & 7.2K &  9.0 & 14.6 & 27.7 &  8.7 &  0.0 & \best{13.8} & \best{14.3} & \best{14.2} \\
    \midrule
    \multicolumn{10}{l}{\textit{Weight-space fine-tuning (best rank of the class-names sweep; full sweep in Appendix~\ref{app:lora_sweep})}}\\
    \lora{} ($r{=}64$)                        & 174.6M & \best{20.2} & \best{19.9} & \best{28.6} &  7.1 & \best{0.7} &  9.6 &  9.4 & \best{14.2} \\
    \midrule
    \midrule
    \multicolumn{10}{l}{\textit{Specialist detector (single-task reference)}}\\
    Grounding DINO~\citep{Liu2023GroundingDetection} (zero-shot)  & 0    & 31.2 &  5.0 & 34.3 & 13.0 &  0.4 &  5.7 & 16.9 & 17.3 \\
    Grounding DINO (10-shot fine-tune)                            & 172M & 40.6 & 32.0 & 44.9 & 41.0 & 35.3 & 30.8 & 26.8 & 35.7 \\
    \bottomrule
  \end{tabular}}
\end{table}

Table~\ref{tab:rf20_categories} compares all methods under 10-shot supervision.
Three observations follow.

First, at any comparable parameter budget, \ours{} dominates.
With $7{,}168$ parameters on average (per-domain $L \in \{1,2,3\}$) it reaches $14.2$ all-domain mAP, $1.7$--$3.4$ mAP above every training-free prompt baseline (closest: DetPO at $12.5$) and above every \lora{} rank except $r{=}64$, which it equals while training $24{,}000\times$ fewer parameters.

Second, the \lora{} rank sweep is itself informative: accuracy increases up to $r{=}64$ before \emph{declining} at $r{=}128$ ($13.6$; Appendix~\ref{app:lora_sweep}), revealing diminishing returns from increasing adaptation capacity. Yet even before this decline, the additional accuracy comes at the hidden cost of forgetting. As shown in Figure~\ref{fig:pareto}, the relative drop in NaturalBench VQA group accuracy compared to the unadapted backbone rises with \lora{} rank, reaching $35\%$ at the accuracy-matching $r{=}64$ and $56\%$ at $r{=}128$. In contrast, \ours{} incurs no forgetting, as the soft prompt leaves the backbone weights entirely unchanged.

Third, discrete prompt optimisation under the same multi-class protocol adds at most $1.2$ mAP over a hand-written prompt: GEPA reaches $11.6$ and DetPO $12.5$, versus $11.3$ for class names with instructions.
Gradient access to a continuous prompt, not prompt optimisation per se, is what closes the remaining gap to fine-tuning ($12.5 \to 14.2$).
The contrast in prompt size makes the same point from the other side: \emph{\ours{} uses one to three learnt tokens, while the optimised discrete prompts run to hundreds or thousands of tokens that are re-processed at every inference call.}

\textbf{Why not a specialist detector?}
Table~\ref{tab:rf20_categories} includes the strongest specialist: Grounding DINO fine-tuned on the same 10 shots reaches $35.7$ mAP, $2.5\times$ any generalist method.
If raw single-task mAP is the only goal, a specialist detector remains the right tool.
Our question is different: practitioners increasingly deploy one generalist VLM for detection, VQA, grounding, and captioning at once, and the relevant cost is what adapting that generalist to a new domain does \emph{to the generalist}.
Figure~\ref{fig:pareto} shows the weight-space answer (up to $56\%$ of its VQA ability) and the prompt-space answer (nothing).
The specialist, which can do none of the other tasks, is the boundary of our claim, not a refutation of it.

\textbf{The medical failure.}
On the Medical category, every generalist method collapses ($\leq 1.0$ mAP for all prompt- and weight-space adapters) while the fine-tuned specialist reaches $35.3$.
Ten shots recover nothing in either adaptation space, consistent with the knowledge being genuinely absent from the backbone rather than badly queried.
Where the model does not already know, neither asking nor light teaching helps; we return to this boundary in \S\ref{sec:discussion}.

\begin{finding} \textbf{Finding.} In the 10-shot regime, prompt-space and weight-space adaptation are accuracy-equivalent: $7{,}168$ prompt parameters on average reach $14.2$ all-domain mAP, equal to the best of six \lora{} ranks ($r{=}64$, $174.6$M parameters), with every other rank strictly below the prompt.
\end{finding}

\subsection{What adaptation costs the generalist}
\label{subsec:forgetting}

A generalist VLM is deployed for more than the domain it was just adapted to.
We measure what each adaptation does to two capabilities the detection task never touches: general VQA and referring-expression grounding.
\ours{} cannot move either number: the backbone is frozen and the domain prompt is simply not supplied off-domain, so its off-domain behaviour is \emph{identical} to the base model by construction.
The question is what \lora{} does.

\textbf{Cross-task VQA (NaturalBench).}
Figure~\ref{fig:pareto} (bottom) plots the relative drop in NaturalBench group accuracy (base model: $37.1$ G-Acc) for every rank of the class-names \lora{} sweep.
The drop grows with rank, $35\%$ relative at $r{=}64$ and $56\%$ at $r{=}128$, and it grows through exactly the rank where detection accuracy peaks: the one \lora{} configuration that matches \ours{}'s accuracy pays a third of the backbone's VQA ability for it.
Detection-only evaluation would never see this cost.

\textbf{Cross-task grounding (RefCOCO).}
Forgetting is not uniform, and honesty requires the other half of the picture.
A \emph{low-rank} adapter ($r{=}4$, single source domain) evaluated on the eight standard RefCOCO splits~\citep{refcoco,refcocog} drifts by $-0.01$ points on average, within run-to-run noise (full table in Appendix~\ref{app:refcoco}).
A gentle \lora{} does not measurably forget grounding.
The correct claim is therefore structural, not absolute: \lora{}'s forgetting is a rank- and scale-dependent \emph{risk} that grows with exactly the configurations practitioners scale to when low ranks underperform, whereas the frozen-backbone prompt carries zero risk at any scale, guaranteed rather than measured.

The deployment consequence is architectural: a single frozen backbone serves \emph{all} domains simultaneously, with each domain requiring only an independent $L{\times}d$ soft-prompt parameter block. Because the backbone remains unchanged, introducing a new domain cannot degrade performance on any existing domain or task. In contrast, weight-space adaptation requires either a separate checkpoint per domain or a continual-learning strategy to offer the same guarantee.

\begin{finding} \textbf{Finding.} \lora{}'s detection gains carry a rank-dependent tax: $35\%$ relative VQA degradation on NaturalBench at the accuracy-matching $r{=}64$, and $56\%$ at $r{=}128$; at $r{=}4$ the RefCOCO drift is $-0.01$ points, so the tax is scale-dependent rather than universal. \ours{} pays zero at any scale, by construction.
\end{finding}

\subsection{A higher-baseline check: LVIS Rare 50}
\label{subsec:lvis}

\rftwenty{} domains are chosen to be far from web pretraining; a fair worry is that soft prompts only help when the backbone is weak.
LVIS Rare 50~\citep{Gare2026DetPO:Detection,lvis} tests the opposite regime: 50 long-tail LVIS classes whose names are ordinary English nouns, where the frozen baseline is already $2\times$ higher ($22.4$ mAP, vs.\ $10.8$ on \rftwenty{}) despite still being far from saturated.

\begin{table}[t]
  \centering
  \caption{\textbf{Gains survive a higher baseline.}
    \ours{} adds $+3.3$ mAP over the $22.4$ frozen baseline ($2\times$ \rftwenty{}'s) at its best length ($L{=}1$); a mis-sized prompt reverses the gain ($L{=}3$ falls to $20.7$, below baseline), the same length-sensitivity \S\ref{subsec:length} finds on \rftwenty{}.
    LVIS Rare 50~\citep{Gare2026DetPO:Detection}, mAP@50:95.}
  \label{tab:lvis_rare50}
  \small
  \setlength{\tabcolsep}{8pt}
  \begin{tabular}{lcc}
    \toprule
    Method & mAP@50:95 & $\Delta$ \\
    \midrule
    \hard{} (frozen baseline) & 22.4 & -- \\
    \ours{} (IST, $L{=}1$)    & \best{25.7} & $+3.3$ \\
    \bottomrule
  \end{tabular}
\end{table}

\ours{} adds $+3.3$ mAP on top of that higher baseline (Table~\ref{tab:lvis_rare50}), comparable to its margin over hard prompts on \rftwenty{} ($+3.4$), rather than overfitting its 10 shots per class.
The gain is length-sensitive in the direction \S\ref{subsec:length} predicts: $L{=}1$ wins, while $L{=}3$ falls to $20.7$, \emph{below} the baseline.
A mis-sized prompt does not merely plateau, it hurts, which is why the per-domain $L$ sweep (with the $L{=}0$ control) is part of the method rather than an optional tuning step.

\subsection{Generalising to a different VLM family: Gemma-4-12B}
\label{subsec:gemma4}

Every result so far trains and evaluates on a single VLM family (\qwen{}); the cross-model study of \S\ref{subsec:transfer} moves a prompt \emph{trained} on \qwen{}-8B to a newer \qwen{} release, still within-family.
A stronger generality claim requires training the method fresh, from scratch, on a genuinely unrelated backbone.
We repeat the recipe -- cross-modal-boundary placement, space-token initialisation, a per-domain hyperparameter search -- on \texttt{Gemma-4-12B-it}~\citep{gemma}, on the 9-dataset ablation pool (\S\ref{subsec:setup}).
Because Gemma-4's optimisation landscape turned out to be noisier than \qwen{}'s (Appendix~\ref{app:gemma4}), the per-domain search here also sweeps format, learning rate, gradient-accumulation, and initialisation, not length alone, all selected on a held-out split of the 10-shot support set, never on test.

\begin{table}[h]
  \centering
  \caption{\textbf{The recipe generalises to a second, unrelated VLM family, but the search's holdout margin does not reliably predict the test-time margin.}
    \texttt{Gemma-4-12B-it}, 9-dataset ablation pool, mAP@50:95.
    \textbf{Test $\Delta$}: the search-selected config's held-out-set-selected hyperparameters, evaluated on test, minus the frozen baseline -- the number a practitioner would actually get by following the search.
    \best{Bold}: dataset where the search-selected config wins on test.
    dentalai is the pool's floor control (\S\ref{subsec:setup}), reported but excluded from the win-count and the mean.
    Full per-dataset holdout-vs-test comparison, search methodology, and every config in Appendix~\ref{app:gemma4}.}
  \label{tab:gemma4}
  \small
  \setlength{\tabcolsep}{6pt}
  \begin{tabular}{lccc}
    \toprule
    Dataset & Baseline & \ours{} (search-selected) & Test $\Delta$ \\
    \midrule
    wb-prova                    &  18.5 & \best{24.3} & $+5.8$ \\
    all-elements                &  13.5 & \best{17.4} & $+3.9$ \\
    new-defects-in-wood         &   3.7 & \best{ 3.9} & $+0.2$ \\
    aerial-airport               &  11.3 & \best{11.4} & $+0.1$ \\
    actions                     &   0.8 & \best{ 1.0} & $+0.2$ \\
    dentalai (floor)             &   0.5 &  0.6 & $+0.1$ \\
    flir-camera-objects         &  14.0 &  13.9 & $-0.1$ \\
    lacrosse-object-detection   &  11.3 &  10.5 & $-0.8$ \\
    defect-detection            &   6.6 &   3.7 & $-2.9$ \\
    \midrule
    \textbf{Mean (above-floor)}  &  10.0 &  \best{10.8} & $+0.8$ \\
    \bottomrule
  \end{tabular}
\end{table}

The recipe transfers: the search-selected soft prompt beats the frozen baseline on 5 of 8 above-floor datasets (Table~\ref{tab:gemma4}), including a $+5.8$ mAP gain on wb-prova and $+3.9$ on all-elements.
But the search's own holdout margin is a poor predictor of that test gain.
Averaged over the same 8 datasets, the selected config's holdout margin over its own control is $+11.0$ mAP; the realised test margin is $+0.8$, roughly an order of magnitude smaller, and on 3 datasets (defect-detection, flir-camera-objects, lacrosse-object-detection) the sign flips outright: a holdout margin of $+4.9$ to $+10.9$ mAP becomes a test \emph{loss}.
defect-detection is the sharpest case: the selected config wins its holdout comparison by $4.9\times$ the control, then loses to baseline by $2.9$ mAP on test, worse than the simplest untuned default we tried for that dataset (Appendix~\ref{app:gemma4}).
This is not the behaviour \S\ref{sec:ablations}'s screening regime shows on \qwen{}, where a short holdout sweep reliably separates configs by a wide, test-consistent margin; on Gemma-4 it does not, at least not with the training budget used here.

\begin{finding} \textbf{Finding.} The soft-prompt recipe generalises to a second, architecturally unrelated VLM family (Gemma-4-12B), beating the frozen baseline on 5 of 8 above-floor datasets by a mean of $+0.8$ mAP. But holdout-based hyperparameter selection, which reliably predicts test performance for \qwen{} (\S\ref{sec:ablations}), does not transfer as a selection procedure to Gemma-4: the mean holdout margin ($+11.0$ mAP) overstates the mean realised test margin by roughly $14\times$, and on 3 of 8 datasets the selected config's holdout win becomes a test loss. Backbone generality and search-transfer reliability are separate claims, and only the first holds here without qualification.
\end{finding}

\section{Ablation Study}
\label{sec:ablations}

The method rests on two claims: position matters and initialisation matters.
This section tests each in isolation, plus the prompt-length choice both interact with.
All ablations run on the fixed 9-dataset pool of \S\ref{subsec:setup}, which covers every RF20 super-category, in the 5-epoch fast-training regime with each dataset's canonical $L$; mAP@50:95, single runs.
Every table carries the frozen baseline as a reference row.

\subsection{Injection position}
\label{subsec:position}

We ablate all six orderings of image (I), soft tokens (S), and text (T) across the pool (Table~\ref{tab:placement}): the two cross-modal bridges, the two soft-last suffixes, and the two soft-first prefixes.

\begin{table}[t]
  \centering
  \caption{\textbf{Where the tokens are injected is worth up to $6\times$ mAP.}
    The cross-modal boundary (IST) wins 6 of 8 above-floor datasets, beats the frozen baseline on all 8, and is never worst; neither soft-first prefix (SIT, STI) ever wins one.
    \best{Bold}: best position per dataset.
    dentalai is the floor control: five of six orderings score exactly $0.0$ and STI alone reaches $0.2$, consistent with the domain-knowledge gap discussed in \S\ref{subsec:main} rather than any one position choice.
    $^{\dagger}$wb-prova's TIS run collapsed to $0.0$.}
  \label{tab:placement}
  \small
  \setlength{\tabcolsep}{4pt}
  \resizebox{\textwidth}{!}{%
  \begin{tabular}{llccccccccc|c}
    \toprule
    Position & Description
      & \makecell{aerial} & \makecell{actions}
      & \makecell{all-\\elements} & \makecell{defect-\\detection}
      & \makecell{dentalai} & \makecell{flir}
      & \makecell{lacrosse} & \makecell{new-\\defects}
      & \makecell{wb-\\prova$^{\dagger}$} & \makecell{Avg} \\
    \midrule
    \multicolumn{2}{l}{Frozen baseline ($L{=}0$)}
                             & 4.4 & 2.9 & 10.6 & 4.9 & 0.0 & 9.3 & 14.2 & 5.9 & 27.6 & 8.9 \\
    \midrule
    IST & Cross-modal bridge & 8.0 & \best{4.0} & 11.7 & \best{8.0} & 0.0 & \best{16.6} & \best{24.0} & \best{11.0} & \best{28.8} & \best{12.4} \\
    TSI & Text-vision bridge & \best{11.4} & 1.9 & \best{14.4} & 6.7 & 0.0 & 5.1 & 4.1 & 8.2 & 26.7 & 8.7 \\
    \midrule
    ITS & Suffix             & 6.3 & 3.3 & 9.1 & 5.6 & 0.0 & 6.8 & 17.1 & 9.1 & 25.9 & 9.2 \\
    TIS & Suffix, modalities swapped & 7.2 & 0.8 & 13.1 & 6.7 & 0.0 & 13.2 & 18.7 & 8.1 & 0.0 & 7.5 \\
    \midrule
    SIT & Prefix (NLP transplant) & 6.3 & 0.8 & 10.4 & 5.6 & 0.0 & 11.8 & 18.3 & 10.0 & 22.8 & 9.6 \\
    STI & Prefix, modalities swapped & 6.7 & 2.0 & 6.7 & 2.8 & \best{0.2} & 10.2 & 6.2 & 5.2 & 22.3 & 6.9 \\
    \bottomrule
  \end{tabular}}
\end{table}

IST wins 6 of the 8 above-floor pool datasets, beats the frozen baseline on all 8, and is never the worst ordering.
IST leads clearly ($12.4$ average); the prefix SIT, the plain suffix ITS, and the other boundary ordering TSI cluster in a middle tier ($9.6$, $9.2$, $8.7$) closer to each other than to either extreme; the mismatched-modality suffix and prefix trail (TIS $7.5$, STI $6.9$).
SIT's respectable average does not translate into ever winning: despite the middle-tier score, neither prefix ordering wins a single above-floor dataset, IST does; on lacrosse the spread between the best and worst ordering is ${\sim}6\times$ ($24.0$ vs.\ $4.1$).
A prompt that precedes all context has nothing to condition on; at the cross-modal boundary, the same tokens see every visual patch and are seen by every text token.
This is the concrete sense in which multimodal soft prompting is not NLP prompt tuning transplanted: the transplant's default placement, SIT, gives up $2.8$ average mAP to the boundary despite its middling average and never wins outright, and reversing the modalities on top of it (STI) gives up nearly a third more ($9.6\to6.9$).

The exceptions are systematic rather than noise, and they polarise.
TSI (text before image) wins exactly the datasets where the text query alone is most informative, single-class aerial ($+58\%$ over IST) and the document-heavy all-elements, and it collapses on lacrosse, a multi-class sports scene ($4.1$ vs.\ IST's $24.0$).
When the query names one known concept, priming the model with text first appears to help; when the relevant class depends on what is in the image, it hurts.
We use IST as the default and treat TSI as the alternative worth testing on single-class or text-dominant domains.

\begin{finding} \textbf{Finding.} Injection position spans $6.9$ to $12.4$ average mAP across the six orderings of image, soft tokens, and text, up to $6\times$ on a single dataset;
the cross-modal boundary placement (IST) achieves the highest performance on 6 of the 8 above-floor pool datasets, while neither soft-first prefix ordering (SIT or STI) wins on any of them. Interestingly, STI is the only ordering to achieve non-zero performance on the challenging dentalai dataset (0.2 vs. 0.0 mAP), suggesting that placing soft prompts before the image can occasionally aid adaptation in particularly difficult domains.
\end{finding}

\subsection{Prompt length}
\label{subsec:length}

\begin{figure}[t]
  \centering
  \includegraphics[width=0.8\textwidth]{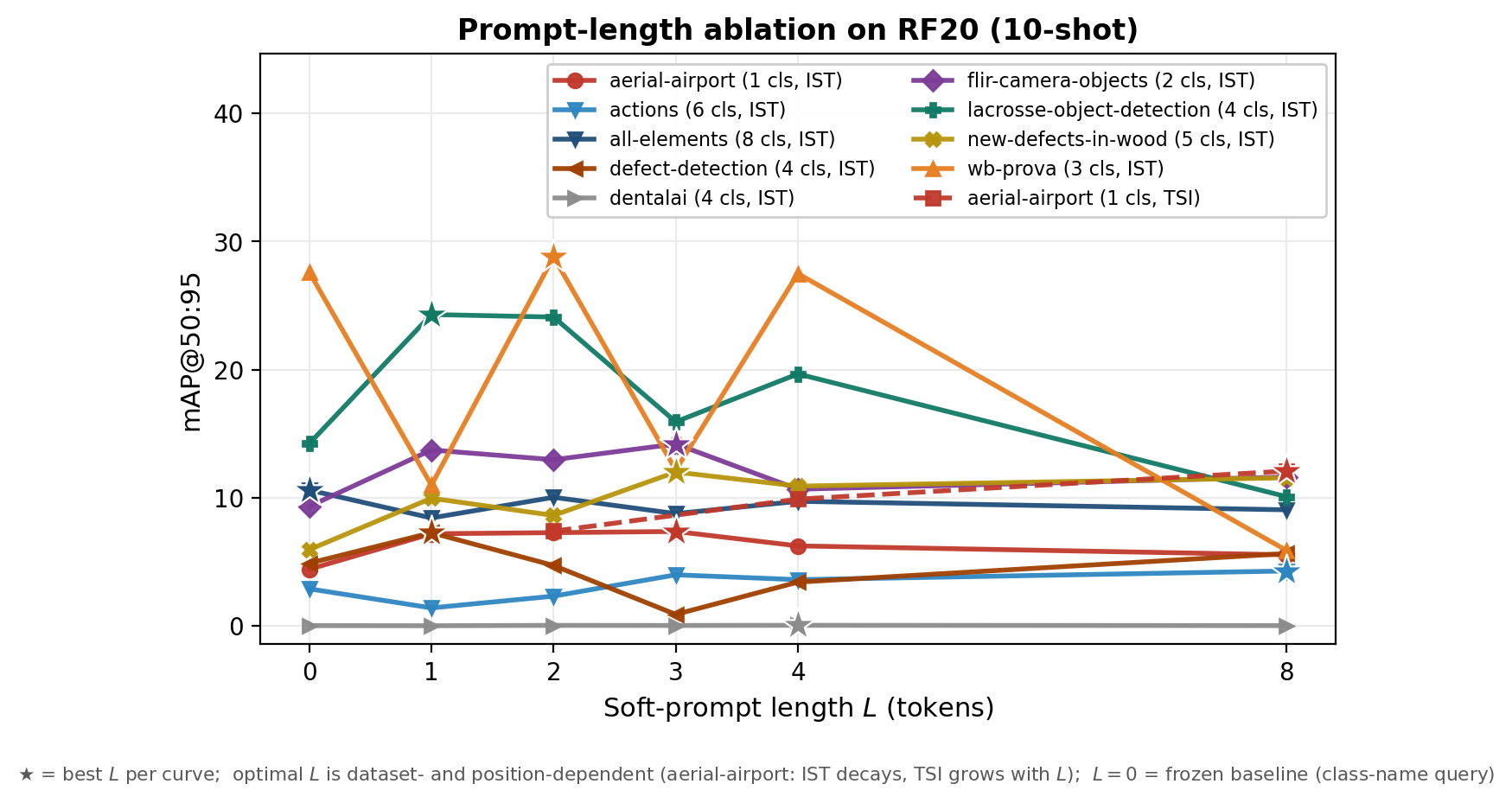}
  \caption{\textbf{Optimal prompt length is small and dataset-dependent, and a mis-sized prompt can lose to $L{=}0$.}
    mAP@50:95 vs.\ $L$ (10-shot, 5-epoch training) on the nine ablation-pool datasets under IST, plus the aerial-airport TSI curve (dashed); stars mark each curve's best $L$, and $L{=}0$ is the frozen baseline (Table~\ref{tab:rf20_categories} source).
    On aerial-airport, IST decays with $L$ while TSI grows, reaching $12.1$ at $L{=}8$: length and position interact.
    dentalai is the flat ${\sim}0$ floor control.}
  \label{fig:prompt_length}
\end{figure}

Figure~\ref{fig:prompt_length} sweeps $L \in \{0,1,2,3,4,8\}$ on all nine pool datasets (IST), plus the aerial-airport TSI curve; the $L{=}0$ points are the frozen baselines of Table~\ref{tab:placement}.

No universal $L$ exists, and no clean class-count rule survives the data: lacrosse peaks at $L{=}1$ ($24.3$) and loses more than half of that by $L{=}8$; flir peaks at $L{=}3$; new-defects drifts upward to $L{=}3$; aerial is flat until $L{=}3$ and then decays.
Two results discipline any tuning recipe.
First, on wb-prova a mis-sized prompt is catastrophic ($5.9$ at $L{=}8$, against a $27.6$ baseline): a soft prompt can hurt, and $L{=}0$ belongs in every sweep as the sanity control.
Second, length interacts with position: on aerial, IST decays with $L$ while TSI (dashed in Figure~\ref{fig:prompt_length}) improves monotonically to $L{=}8$ ($12.1$, the best aerial result in our sweeps).
The practical recipe is a short sweep over $\{0,1,2,3,4,8\}$ per domain; each configuration is a single-GPU run whose training loop takes minutes on low-resolution domains and a median of ${\sim}2$ hours across our ablation runs (Appendix~\ref{app:impl}).

\subsection{Initialisation}
\label{subsec:init}

Table~\ref{tab:init} compares five initialisations across the pool.

\begin{table}[t]
  \centering
  \caption{\textbf{What should soft prompts be initialised with? The token that means nothing.}
    A single space token wins 5 of 8 above-floor datasets and beats the frozen baseline on all 8; the semantically richer initialisations are the least reliable ones (task-text collapses on lacrosse, $6.6$ vs.\ space's $24.1$; class-seeded collapses on all-elements, $5.3$ vs.\ $11.6$).
    \best{Bold}: best initialisation per dataset.
    dentalai is the floor control: every initialisation sits at the baseline, consistent with the domain-knowledge gap discussed in \S\ref{subsec:main} rather than any one initialisation choice.}
  \label{tab:init}
  \small
  \setlength{\tabcolsep}{4pt}
  \resizebox{\textwidth}{!}{%
  \begin{tabular}{llccccccccc|c}
    \toprule
    Init strategy & Description
      & \makecell{aerial} & \makecell{actions}
      & \makecell{all-\\elements} & \makecell{defect-\\detection}
      & \makecell{dentalai} & \makecell{flir}
      & \makecell{lacrosse} & \makecell{new-\\defects}
      & \makecell{wb-\\prova$^{\dagger}$} & \makecell{Avg} \\
    \midrule
    \multicolumn{2}{l}{Frozen baseline ($L{=}0$)}
                                  & 4.4 & 2.9 & 10.6 & 4.9 & 0.0 & 9.3 & 14.2 & 5.9 & 27.6 & 8.9 \\
    \midrule
    Space (\texttt{" "})          & Single whitespace token & 8.0 & \best{3.9} & \best{11.6} & \best{8.0} & 0.0 & \best{16.6} & \best{24.1} & 11.0 & 28.8 & \best{12.4} \\
    Mean                          & Mean vocab embedding    & \best{9.2} & \best{3.9} & 6.7 & 4.9 & 0.0 & 7.5 & 14.4 & 7.4 & 13.3 & 7.5 \\
    Xavier                        & Xavier uniform          & 6.7 & 3.0 & 8.4 & 5.0 & 0.0 & 12.1 & 21.2 & \best{13.1} & 19.5 & 9.9 \\
    Class-seeded                  & Class name tokens       & 7.5 & 2.2 & 5.3 & 4.7 & 0.0 & 13.3 & 13.6 & 12.2 & 22.6 & 9.0 \\
    Task-text (\texttt{"detect"}) & Task description tokens & 7.6 & 3.6 & 7.3 & 6.5 & 0.0 & 13.3 & 6.6 & 10.8 & \best{30.3} & 9.6 \\
    \bottomrule
  \end{tabular}}
\end{table}

The single space token wins 5 of the 8 above-floor pool datasets, has the best average ($12.4$), and beats the frozen baseline on all 8.
The semantically richer initialisations are the less reliable ones: task-text (``detect'') collapses on lacrosse ($6.6$ vs.\ space's $24.1$), and class-seeded starts collapse on all-elements ($5.3$ vs.\ $11.6$).
This is the identity principle of \S\ref{subsec:method_init} in the data: an initialisation that preserves the prompt's meaning starts from the frozen baseline's behaviour and can only improve, while a semantic start first has to undo its own interference.
The one dataset where space is not the best initialisation is wb-prova$^{\dagger}$: task-text wins by a narrow margin ($30.3$ vs.\ space's $28.8$), consistent with a dataset whose query, not its visuals, is the binding constraint.
Mean-embedding initialisation offers a cautionary aside: it achieves the lowest validation \emph{loss} on every dataset while winning or tying on mAP for only two (aerial outright, actions tied with space).
Across all three ablations, validation loss does not track detection mAP; practitioners should select models on a task metric, not the language-model loss.

\begin{finding} \textbf{Finding.} A single space token is the best initialisation, winning 5 of 8 above-floor pool datasets ($12.4$ avg vs.\ $7.5$ for mean-embedding and $9.6$ for task-text): initialisations that preserve the prompt's meaning dominate initialisations that encode task semantics.
\end{finding}

\section{Prompt Distillation: What Do the Tokens Encode?}
\label{sec:analysis}

The ablations say success rides on prompt-like choices: position, brevity, meaning preservation.
If that is right, the learnt tokens should behave like language in other respects too.
This section reads them out directly, verbalises them, and transfers them.

\subsection{Two regimes: class surrogates vs.\ general instructions}
\label{subsec:retrieval}

\begin{figure}[t]
    \centering
    \includegraphics[width=\textwidth]{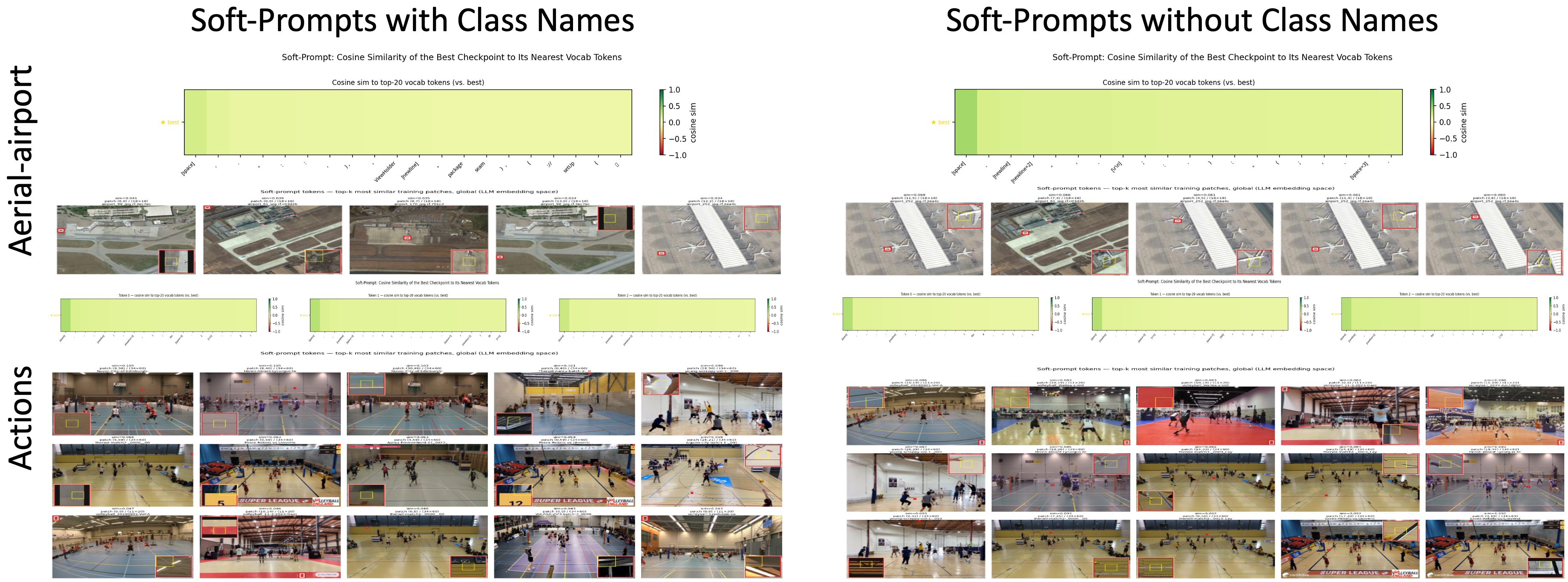}
    \caption{
      \textbf{What the tokens learn depends on what the prompt lacks.}
      Nearest text tokens and image patches for trained soft prompts. When the class name is withheld from the query (probe configuration), the soft tokens' nearest neighbours become image patches of the target class, effectively acting as a visual surrogate for the missing class name. When class names are present (IST/TSI), the tokens no longer align with object patches and instead seem to encode general detection instructions. In both settings, the nearest text tokens remain close to those at initialization, indicating that the learnt embeddings stay centered around their initialization despite adaptation. Remarkably, when the model must represent a missing class concept, it does so more naturally in the visual embedding space than in the linguistic one, suggesting that visual features provide a more effective representation of object identity than nearby words.
      }
    \label{fig:retrival}
\end{figure}

To read the tokens, we retrieve their nearest neighbours in the model's own representation spaces: the text-vocabulary embeddings and a cache of visual patch embeddings from the frozen encoder (Figure~\ref{fig:retrival}).
The answer depends on what the prompt was \emph{forced} to learn.
In a probe configuration where the class name is removed from the text query and the soft tokens must stand in for it, the trained tokens retrieve the object itself: their nearest visual patches are patches of the detected class.
In the high-performing IST/TSI configuration, where the class names remain in the query, the trained tokens do \emph{not} match object patches; describing the class again would be redundant, and the tokens instead act as general task instructions.

The nearest neighbours are not just interpretable, they are functional: substituting each trained token with its nearest text token (or visual patch) at evaluation retains most of the adaptation on the ablation pool (Appendix~\ref{app:substitution}).

\subsection{Verbalising the prompt: \distill{}}
\label{subsec:distill}

With the learnt soft-tokens living in the model's own representational space, a VLM should be able to verbalize the meaning of soft tokens.
We find that Naively asking the model to describe the learnt soft tokens leads to the model regarding the soft-token to its closest text tokens which is usually the initialized (space) token.
We find that the soft-tokens carry meaning in the context that they where learnt with i.e. with the Image and Prompt Query, under this setting attempts to reexplain the soft-tokens fails as the soft-tokens focus the model to predict detections.
Thus our solution is to ask the model to describe the image first before predicting bounding boxes thus we can compare the descriptions generated by this pipeline with the control where no soft-tokens are presented, this should show the qualitative deference in object descriptions which we can use as a prompt for detection to get a quantitative measure.

\begin{figure}[t]
    \centering
    \includegraphics[width=0.75\textwidth]{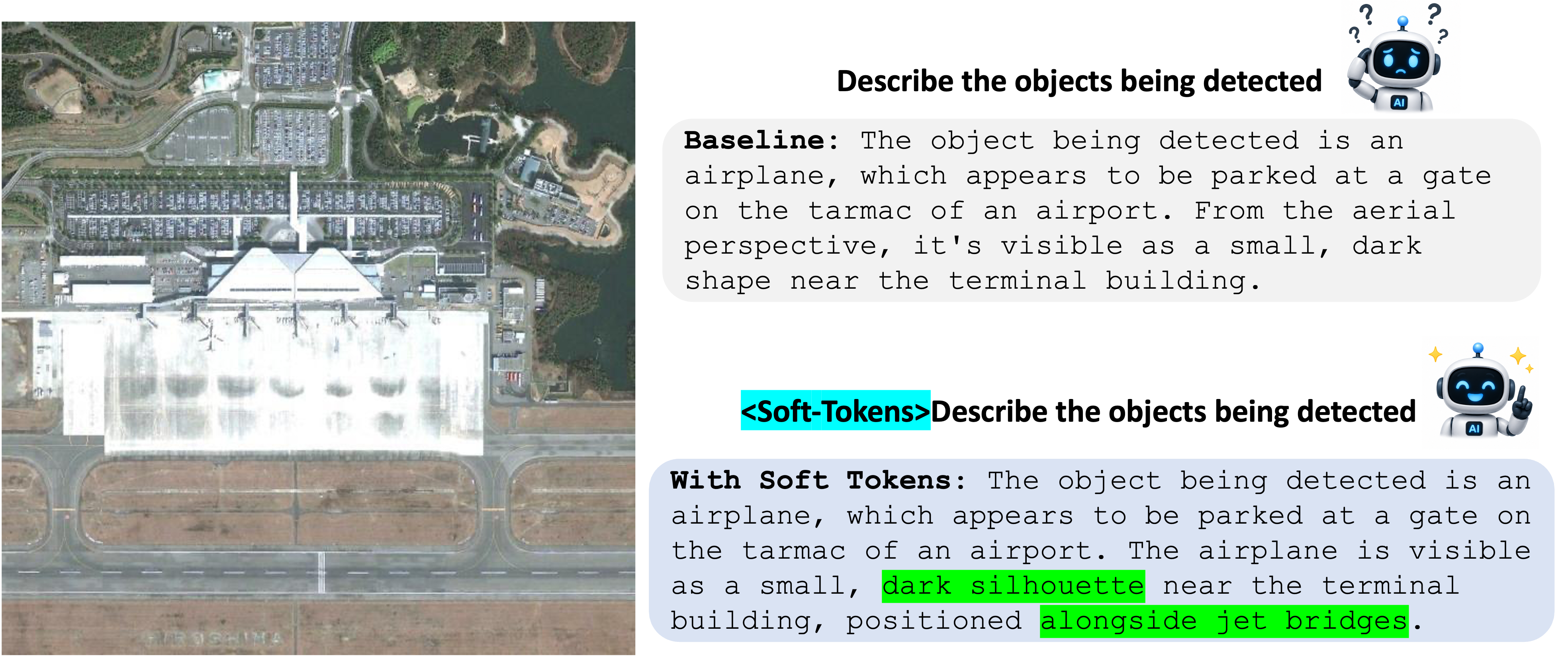}
    \caption{\textbf{Descriptions generated under the soft prompt notice finer object detail.}
      Asked to describe the objects being detected, the model with soft tokens active names distinguishing features that the base model's description omits; these richer descriptions are what stage 2 of \distill{} summarises into the distilled prompt.}
    \label{fig:distill}
\end{figure}

If the learned token embeddings capture prompt-like semantics, those semantics should be recoverable as text, and the recovered text should retain predictive utility.
Table~\ref{tab:distillation} evaluates the generate-summarise-evaluate recipe of \S\ref{subsec:method_distill} and contains two honest decompositions.
First, the distilled prompt is competitive with purpose-built discrete prompt search: $12.5$ all-domain mAP, a tie with DetPO ($12.5$) and above GEPA ($11.6$), despite being obtained indirectly (by verbalising a gradient-trained object) rather than by searching prompt space.
Second, most of that margin comes from the describe-then-summarise recipe itself: the base-source variant already reaches $12.2$, and the soft prompt adds $+0.3$ on average.
The soft prompt's contribution is real but concentrated where the frozen backbone is weakest, Aerial ($+4.1$) and Documents ($+2.6$), and neutral-to-negative on categories the base model already describes well.
Qualitatively (Figure~\ref{fig:distill}), descriptions generated under the soft prompt highlight finer distinguishing features of the target objects, which is the plausible carrier of the gain.
The distilled prompt recovers $12.5$ of the $14.7$ ceiling set by the per-dataset-best soft prompt (test-split position selection; \S\ref{subsec:setup}, disclosure ii) while being editable, auditable, and usable with \emph{any} model that accepts text, including closed-source APIs that expose no embedding interface.

\begin{table}[t]
  \centering
  \caption{\textbf{Verbalised soft prompts match dedicated prompt search, and most of the recipe's value is the recipe -- but the marginal $+0.3$ average hides a real, dataset-limited gain.}
    Distilled global hard-prompt test mAP@50:95 per super-category; $12.2$ of \distill{}'s $12.5$ comes from the describe-summarise recipe alone, and only $+0.3$ from the soft prompt on average, concentrated where the backbone is weakest (Aerial $+4.1$, Documents $+2.6$).
    \distill{} (soft): descriptions generated with the soft prompt active; base recipe: the identical pipeline without soft tokens; $\Delta$ isolates the soft prompt's contribution.
    \ours{} row: the trained soft prompt's own test mAP, the ceiling distillation aims to recover.
    \best{Bold}: best method using no soft tokens at inference (a tie between \distill{} and DetPO).
    \textbf{Gain}/\textbf{Loss}: mean per-dataset $\Delta$ (soft $-$ base), among the 20 \rftwenty{} datasets where the soft prompt helped/hurt the recipe, with the dataset count in parentheses.}
  \label{tab:distillation}
  \small
  \setlength{\tabcolsep}{4pt}
  \resizebox{\textwidth}{!}{%
  \begin{tabular}{lccccccc|cc|c}
    \toprule
    \textbf{Method} & \textbf{Aerial} & \textbf{Doc.} & \textbf{F.\,\&\,F.}
                    & \textbf{Ind.} & \textbf{Med.} & \textbf{Sports}
                    & \textbf{Other} & \textbf{All} & \textbf{Gain ($n$)} & \textbf{Loss ($n$)} \\
    \midrule
    \ours{} (soft prompt, ceiling)         & 12.2 & 15.1 & 28.5 & 8.8 & 0.0 & 13.8 & 14.3 & 14.7 & & \\
    \midrule
    \distill{} (soft source, ours)         & 9.6 & 10.8 & 26.3 & 9.2 & 0.1 & 10.9 & 10.9 & \best{12.5} & & \\
    Base recipe (no soft tokens)           & 5.5 &  8.2 & 27.0 & 8.5 & 0.0 & 11.4 & 12.1 & 12.2 & & \\
    $\Delta$ (soft $-$ base)               & $+4.1$ & $+2.6$ & $-0.7$ & $+0.7$ & $+0.1$ & $-0.5$ & $-1.2$ & $+0.3$ & $+1.9$ (10) & $-1.4$ (10) \\
    \midrule
    GEPA~\citep{AgrawalGEPA:LEARNING}      & 6.3 & 10.8 & 22.4 & 8.5 & 0.1 & 12.8 & 11.5 & 11.6 & & \\
    DetPO~\citep{Gare2026DetPO:Detection}  & 8.6 & 10.1 & 26.7 & 9.9 & 0.1 & 11.0 & 10.5 & \best{12.5} & & \\
    \bottomrule
  \end{tabular}}
\end{table}

\begin{finding} \textbf{Finding.} A soft prompt can be verbalised: the distilled natural-language prompt reaches $12.5$ all-domain mAP, a tie with the strongest dedicated prompt search (DetPO, $12.5$) and above GEPA ($11.6$); $12.2$ of that comes from the describe-summarise recipe and $+0.3$ from the soft prompt, concentrated where the backbone is weak (Aerial $+4.1$). That $+0.3$ average is a blend, not a uniform effect: the soft prompt helps the recipe on 10 of 20 \rftwenty{} datasets (mean $+1.9$ mAP) and hurts it on the other 10 (mean $-1.4$), so the gain is real but dataset-limited, not a broad, small lift everywhere.
\end{finding}

Appendix~\ref{app:synth_arith} isolates this same soft-to-hard behaviour from the vision pipeline entirely, in a minimal synthetic-arithmetic setting: a trained soft prompt is uninterpretable when queried alone, but becomes reliably, often exactly, describable once the query preserves the context it was trained in.

\subsection{Transfer across model versions}
\label{subsec:transfer}

Unlike weight updates, a trained soft prompt is an input, so it can at least be \emph{offered} to a different model.
We test the simplest protocol: prompts trained on \qwen{}-8B are loaded, with no projection or retraining, into Qwen3.5-9B (Table~\ref{tab:transfer}).

\begin{table}[t]
  \centering
  \caption{\textbf{Naive cross-model transfer recovers about $17\%$ of the adaptation gap on average -- but that average blends a real gain on most datasets against a real loss on the rest.}
    The transferred prompt nets $+0.8$ mAP over the 9B zero-shot baseline, only $0.8$ of the full $+4.8$ mAP gap to the 8B origin, and sits $4.0$ mAP below it; per-category, it helps most on Sports ($+5.3$) and hurts on Industrial ($-2.9$, Appendix~\ref{app:transfer}).
    \qwen{}-8B $\to$ Qwen3.5-9B, all-domain mAP@50:95 over the 20 \rftwenty{} datasets.
    9B baseline: target model, no prompt; 9B + transfer: the 8B-trained prompt loaded as-is; 8B origin: the same prompt on its training model (upper bound).
    \textbf{Gain}/\textbf{Loss}: mean per-dataset $\Delta$ vs.\ the 9B baseline, among the datasets where transfer helped/hurt, with the dataset count in parentheses.}
  \label{tab:transfer}
  \small
  \setlength{\tabcolsep}{6pt}
  \begin{tabular}{lcccc}
    \toprule
    Method & All-domain mAP & $\Delta$ & Gain ($n$) & Loss ($n$) \\
    \midrule
    Qwen3.5-9B baseline       & 9.9  & --     &            &            \\
    9B + transferred prompt   & 10.7 & $+0.8$ & $+2.9$ (13) & $-3.1$ (7) \\
    8B origin (upper bound)   & 14.7 & $+4.8$ &            &            \\
    \bottomrule
  \end{tabular}
\end{table}

The transferred prompt nets $+0.8$ all-domain mAP over the 9B zero-shot baseline, helping most on the Sports ($+5.3$) and Documents ($+3.3$) categories and hurting on Industrial ($-2.9$); per-category numbers are in Appendix~\ref{app:transfer}.
That $+0.8$ average is again a blend, not a uniform effect: transfer beats the 9B baseline on 13 of the 20 \rftwenty{} datasets, by a mean of $+2.9$ mAP, and loses on the other 7, by a mean of $-3.1$; the two groups are similar in size and opposite in sign, so the small headline average is the arithmetic of a real split, not a small effect everywhere.
It recovers only part of the adaptation, sitting $4.0$ mAP below its 8B-origin performance.
So the tokens carry some model-independent domain signal, as a text prompt would, but most of their value is expressed in the origin model's embedding geometry; a learnt projection between embedding spaces is the natural next step, as in text-only prompt transfer~\citep{vu2022spot}.

\begin{finding} \textbf{Finding.} Cross-model transfer's $+0.8$ mAP average is a blend of a real gain and a real loss, not a small uniform effect: it beats the 9B zero-shot baseline on 13 of 20 \rftwenty{} datasets (mean $+2.9$ mAP) and loses on the other 7 (mean $-3.1$). The gain is genuine where it appears, but is not a property of the transfer protocol uniformly across domains.
\end{finding}

\textbf{The embeddings transfer better than their own description.}
\S\ref{subsec:distill} shows a trained prompt can be verbalised into a natural-language instruction on its own model.
A natural question is whether that verbalised instruction is a \emph{better} cross-model transfer vehicle than the raw embeddings above, since it is plain text and never touches the target model's embedding space at all: we distil each dataset's canonical \qwen{}-8B soft prompt into a hard prompt (the \S\ref{subsec:distill} describe-summarise recipe, best of its global/per-class variant) and evaluate it directly on Qwen3.5-9B with no soft tokens, against the same raw-embedding transfer as Table~\ref{tab:transfer} (Table~\ref{tab:distill_transfer}; per-dataset results in Appendix~\ref{app:distill_transfer}).

\begin{table}[t]
  \centering
  \caption{\textbf{Raw-embedding transfer beats its own distilled description on 18 of 20 datasets.}
    Two cross-model transfer strategies for the same \qwen{}-8B-trained soft prompt, evaluated on Qwen3.5-9B: loading the trained embeddings directly (as in Table~\ref{tab:transfer}) vs.\ evaluating the \S\ref{subsec:distill} hard prompt distilled from that same prompt, with no soft tokens at all.
    All-dataset mean mAP@50:95 over the 20 \rftwenty{} datasets (per-dataset results in Appendix~\ref{app:distill_transfer});``distilled'' takes the better of the global/per-class variant per dataset.}
  \label{tab:distill_transfer}
  \small
  \setlength{\tabcolsep}{8pt}
  \begin{tabular}{lccc}
    \toprule
    Transfer strategy & All-dataset mAP & Datasets won & Median mAP \\
    \midrule
    Distilled hard prompt (best variant) & 9.3  & 2/20  & 5.8 \\
    Raw embeddings (direct)              & 11.6 & 18/20 & 9.4 \\
    \bottomrule
  \end{tabular}
\end{table}

Direct embedding transfer wins on 18 of the 20 datasets, by a mean of $+2.3$ mAP and up to $+9.9$ (lacrosse-object-detection, $7.1\to17.1$).
The two exceptions are both near-floor swaps below $4$ mAP either way (defect-detection, $3.2\to3.6$; gwhd2021, $0.9\to1.8$), where any ordering is noise, not signal.
Verbalising a soft prompt is useful for interpretability and for search-space competitiveness (\S\ref{subsec:distill}), but it is lossy as a transfer channel: the continuous vector, even off its training model, carries more of the original prompt's signal than the paragraph the model itself would write to describe it, which matches the intuition that natural language is a lower-bandwidth bottleneck than the embedding it was distilled from.

\begin{finding} \textbf{Finding.} The trained embeddings transfer across model versions better than their own natural-language description: raw-embedding transfer beats the distilled hard prompt on 18 of 20 \rftwenty{} datasets (mean $+2.3$ mAP, up to $+9.9$), and both exceptions are near-floor swaps below $4$ mAP. A soft prompt is stronger transferable evidence than the text it can be turned into.
\end{finding}

\section{Beyond Detection: Prompting a Robot Policy}
\label{sec:robotics}

Is the recipe specific to detection, or does it adapt any frozen generative model with a multimodal input stream?
We attempt to answer this question by applying our approach to optimizing a vision-language-\emph{action} (VLA) policy: $\pi_{0.5}$~\citep{black2024pi0,intelligence2025pi05}, a flow-matching policy built on a PaliGemma VLM backbone with a separate action-expert module that produces continuous robot actions.

\textbf{Setup.}
We use three atomic tasks from the RoboCasa kitchen suite~\citep{nasiriany2024robocasa} on the PandaOmron mobile manipulator, an embodiment of the base policy was pretrained on with RoboCasa data~\citep{mandlekar2023mimicgen}.
The frozen base policy performs poorly on two of the tasks (\texttt{PickPlaceCounterToCabinet} and \texttt{TurnOnSinkFaucet}, $5.0\%$ zero-shot success each); the third it already partly solves (\texttt{OpenDrawer}, $30.0\%$).
Adaptation uses $K{=}10$ demonstration episodes per task, the manipulation analogue of our 10-shot detection protocol.
Evaluation is closed-loop success rate over 20 freshly sampled scene instantiations (held-out object placements and layouts) per seed, 3 seeds per arm; every arm sees byte-identical scenes via per-episode environment seeding.
The soft prompt follows the detection recipe exactly: IST format, space-token initialisation, trained with the policy frozen.
The \lora{} baselines follow the openpi reference configuration (rank 16); details in Appendix~\ref{app:impl}.

\begin{table}[t]
  \centering
  \caption{\textbf{The placement lever generalises to VLAs, and base competence marks its boundary.}
    A dual prompt reaching the action expert lifts the two near-floor tasks from $5.0\%$ to $23.3\%$ and $31.7\%$ success, matching or beating coverage-matched \lora{} at ${\sim}400\times$ fewer parameters; the same prompt \emph{drops} the task the base already partly solves, OpenDrawer, from $30.0\%$ to $21.7\%$ unless early-stopped ($45.0\%$), while every \lora{} variant improves it.
    Prefix-only prompts condition only the PaliGemma backbone; the dual prompt adds $M{=}16$ tokens on the action-expert input, in the gradient path of the flow-matching loss; the early-stopped variant halts training at 750 of 2{,}000 steps.
    Trainable-parameter counts are exact, not estimated: $N{=}8$ prefix tokens at PaliGemma's width $2048$ gives $16{,}384$; the dual prompt adds $M{=}16$ action-expert tokens at width $1024$, giving $16{\cdot}2048 + 16{\cdot}1024 = 49{,}152$ (verified directly against the saved adapter checkpoints' tensor shapes).
    The action-expert tokens are randomly initialised, not space-token initialised like the PaliGemma-side tokens: the action expert has no token vocabulary to seed from.
    \best{Bold}: best prompt-based and best weight-based adapter per task; ``--'': arm not run.}
  \label{tab:robocasa}
  \small
  \setlength{\tabcolsep}{5pt}
  \begin{tabular}{lcccc}
    \toprule
    & & \multicolumn{2}{c}{Weak base} & Competent base \\
    \cmidrule(lr){3-4}\cmidrule(lr){5-5}
    Adapter & Trainable & PickPlace & Faucet & OpenDrawer \\
    \midrule
    $\pi_{0.5}$ base (zero-shot)                  & --     & 5.0\% & 5.0\% & 30.0\% \\
    \midrule
    \multicolumn{5}{l}{\textit{Prompt-space (frozen policy, ours)}}\\
    Soft prompt, prefix-only ($N{=}8$)            & 16K    & 1.7\% & -- & -- \\
    Soft prompt, dual ($N{=}16{+}M{=}16$)         & 49K    & \best{23.3\%} & \best{31.7\%} & 21.7\% \\
    Soft prompt, dual, early-stopped              & 49K    & -- & -- & \best{45.0\%} \\
    \midrule
    \multicolumn{5}{l}{\textit{Weight-space (\lora{}, rank 16)}}\\
    \lora{}, PaliGemma-only (coverage-matched)    & 19.6M  & 23.3\% & 21.7\% & 66.7\% \\
    \lora{}, + action expert                      & 26.5M  & \best{26.7\%} & \best{26.7\%} & \best{78.3\%} \\
    \bottomrule
  \end{tabular}
\end{table}

\textbf{Placement decides whether the prompt can act.}
On \texttt{PickPlaceCounterToCabinet}, a prefix-only soft prompt, the direct transplant of the detection recipe onto the PaliGemma prefix, fails: $1.7\%$ success, at the zero-shot floor, and no sweep of token count, learning rate, or training steps lifts it reliably.
The failure is one of placement, not capacity.
In $\pi_{0.5}$, the training loss is a flow-matching regression on the action-expert output; the PaliGemma prefix output is discarded, so a prefix prompt influences the action only indirectly, through the action expert's attention back into the prefix.
This is our gradient-path argument (\S\ref{subsec:method_training}) taken to its limit: the prompt is no longer merely a weak lever, it is outside the output's direct path.
Moving the prompt fixes it.
A \emph{dual} prompt, $N{=}16$ tokens on the PaliGemma prefix plus $M{=}16$ tokens on the action-expert input, sits directly in the gradient path of the action loss and reaches $23.3\%$ success, matching the coverage-matched PaliGemma-only \lora{} at $400\times$ fewer trainable parameters.
The prefix tokens keep the space-token initialisation of \S\ref{subsec:method_init}; the action-expert tokens are initialised randomly instead, since the action expert has no token vocabulary to seed an identity-preserving start from.
One optimisation detail matters: the flow-matching gradient is high-variance (fresh time step and noise per sample), and a small prompt starves at small batch; gradient accumulation to an effective batch of 128 with a high learning rate is what makes the dual prompt trainable.

\textbf{Where the base is weak, the prompt keeps pace with the weight update.}
The same dual recipe, hyperparameters unchanged, transfers to \texttt{TurnOnSinkFaucet}: from the same $5.0\%$ floor it reaches $31.7\%$, ahead of both the coverage-matched \lora{} ($21.7\%$) and the rank-16 \lora{} that also adapts the action expert ($26.7\%$).

\textbf{If the base policy is competent, the prompt harms perfomance.}
\texttt{OpenDrawer} inverts the picture: the frozen base already succeeds at $30.0\%$, the same full-strength dual prompt drops reduces performance to $21.7\%$, and every \lora{} variant improves it ($66.7$ to $78.3\%$).
The prompt is not failing to learn; it fits the ten demonstrations as well as on the other tasks (train loss $0.02$--$0.05$).
It is a learned, fixed edit of the input that shifts the frozen policy globally, and on a task the base already solves, that shift replaces a competent pretrained behaviour with an overfit ten-demonstration one.
A weight update does not share this failure mode: its per-layer corrections can leave the base computation in place except where the demonstrations demand change.
Restraining the prompt recovers the sign: stopping the same configuration at 750 of 2{,}000 steps yields $45.0\%$, above the base, though still far below rank-16 \lora{} ($78.3\%$).
Read together, the three tasks say that a soft prompt acts as a few-shot \emph{override} of a frozen policy: decisive where the base is near the floor, harmful at full strength where the base is already competent.
This is the action-space analogue of the regime our detection results live in, where the frozen baseline is weak on the adapted domains.

\textbf{Caveats.}
These are three atomic tasks with 20 trials per seed (${\pm}5\%$ per point); the dual prompt's seed spread on PickPlace is $10$--$30\%$; the early-stop point was selected on a seed-0 sweep before its 3-seed re-run.
We read this as a mechanism result, not a benchmark claim.

\begin{finding} \textbf{Finding.} The placement principle carries beyond detection, and base competence marks its boundary: a dual prompt reaching the action expert lifts two near-floor RoboCasa tasks from $5.0\%$ to $23.3\%$ and $31.7\%$ success, matching or beating coverage-matched \lora{} at ${\sim}400\times$ fewer parameters, yet drops an already-competent task from $30.0\%$ to $21.7\%$ unless early-stopped ($45.0\%$). A soft prompt is a few-shot override: it must sit in the gradient path of the output it steers, and it pays off where the frozen model is weak.
\end{finding}

\section{Discussion}
\label{sec:discussion}

\textbf{Why does asking match teaching at 10 shots?}
Ten images provide on the order of tens of labelled boxes: enough to identify \emph{what to attend to}, not enough to estimate millions of weight updates.
\lora{}'s own rank sweep is the cleanest evidence: accuracy peaks at $r{=}64$ and falls at $r{=}128$ (Appendix~\ref{app:lora_sweep}), classic capacity overshoot, while the forgetting cost keeps rising.
The soft prompt puts its ${\sim}7$K parameters entirely inside the frozen model's input distribution, where the backbone's own inductive biases regularise them.
We state this as an interpretation of the observed regime, not a law: with hundreds of shots the balance may move toward weight space, and we have not measured where.
The robotics results draw the same boundary from the other side: when the target behaviour is genuinely new to the module that produces the output, a prompt succeeds only if it reaches that module, and a full fine-tune still leads.
They also expose the boundary's second face: on a task the frozen policy already performs, the prompt's global input edit overwrites competence that a weight update's targeted corrections preserve, and only a restrained (early-stopped) prompt stays above the base.

\textbf{Adaptation is insufficient if the backbone lacks the required knowledge.}
The Medical category is the thesis's own counterexample, and we keep it in the headline table: every prompt- and weight-space adapter on the generalist backbone scores $\leq 1.0$ mAP while a fine-tuned specialist reaches $35.3$.
Ten-shot adaptation, in any space, cannot conjure absent knowledge.
The recommendation that follows is diagnostic: if a short soft-prompt sweep (single-GPU runs; Appendix~\ref{app:impl} timings) does not move a domain off the floor, the failure is knowledge, not phrasing, and a specialist or heavier adaptation is the right tool.

\textbf{Practical recipe.}
For a new domain: IST position; $L$ swept over $\{0,1,2,3,4,8\}$ (TSI worth testing for single-class or text-dominant domains); space-token initialisation; the minimal class-name query the prompt was trained with; model selection by task metric rather than loss.
For a VLA policy: prompt both the VLM prefix and the action-expert input, and train with a large effective batch.
Total cost is a handful of single-GPU runs per domain (median ${\sim}2$ GPU-hours each in our 5-epoch regime; Appendix~\ref{app:impl}).

\textbf{Limitations.}
(1)~\emph{Statistical support:} all numbers are single runs in a regime with $20$--$28\%$ seed spread; per-dataset deltas under ${\sim}1$ mAP are ties, and the headline claim is parity with the best \lora{}, not victory over it.
(2)~\emph{Scope:} the main detection results, the ablations, and the cross-model \emph{transfer} study are all within the \qwen{} family; a from-scratch check on a second, unrelated backbone (Gemma-4-12B, \S\ref{subsec:gemma4}) shows the recipe itself generalises, but also that its per-domain hyperparameter search does not carry \qwen{}'s holdout-to-test reliability to that backbone (Appendix~\ref{app:gemma4}) -- one shot count (10) and three atomic VLA tasks for robotics remain untested outside those settings, and claims are scoped accordingly.
(3)~\emph{Selection:} the per-dataset best-of-IST/TSI checkpoints used as the distillation ceiling and transfer origin were selected on the test split, disclosed in \S\ref{subsec:setup}; the main comparison uses the fixed IST position only.
(4)~\emph{Training memory:} the trainable-parameter count does not translate into proportional training-memory savings; peak memory is dominated by activations of the frozen backbone, and in our measurements soft-prompt training at batch size 4 has peak memory comparable to a \lora{} $r{=}16$ run at batch size 1.
Gradient checkpointing~\citep{gradckpt} is the practical mitigation; storage and deployment efficiency ($L{\times}d$ floats per domain) are unaffected: a trained prompt is $17.6$--$49.6$\,KB on disk depending on $L$, against \lora{} $r{=}64$'s $666$\,MB, a ${\sim}23{,}000\times$ reduction measured directly from the checkpoint files (Appendix~\ref{app:vllm}).
(5)~\emph{Scope of the query:} all \ours{} results use the minimal class-name query; combining trained soft tokens with descriptive per-class instructions, the setting in which \lora{} reaches $14.8$ (Appendix~\ref{app:lora_sweep}), is left to future work.

\section{Conclusion}
\label{sec:conclusion}

We revisited soft prompting for few-shot adaptation of vision-language models and found that two choices decide whether it works: injecting the tokens at the cross-modal boundary, not as a NLP-standard prefix, and 
and initializing them so that the prompt's semantic meaning remains unchanged at the start of training.
With those choices, one to three learnt tokens per domain equal the best of a \lora{} rank sweep on \rftwenty{}, at over $20{,}000\times$ fewer trainable parameters and zero forgetting, while the accuracy-matching \lora{} rank forgets $35\%$ of the backbone's NaturalBench VQA ability.
The learnt tokens keep the virtues of language: they transfer across model versions, and they can be verbalised into editable prompts that match dedicated prompt search.
The same placement principle carries to a frozen robot policy, where a prompt succeeds exactly when it reaches the module that produces the action.
We take these results as evidence that modern VLMs already encode much of what specialised domains require.
Before teaching a model with weight updates, and paying what that costs the generalist, it is worth learning to ask.

\subsubsection*{Ethics Statement}
This work aims to make few-shot adaptation of large vision-language models cheaper and safer: prompt-space adaptation lowers the annotation and compute barrier for specialised domains (environmental monitoring, industrial inspection, medical triage) and, unlike weight updates, cannot silently degrade the deployed model's other capabilities.
Distilled prompts are readable text, a modest transparency benefit over opaque weight deltas.
Risks mirror those of object detection generally, including surveillance applications; we do not identify consequences specific to this method beyond them.

\subsubsection*{Reproducibility Statement}
All benchmarks are public (Roboflow-100-VL, LVIS, NaturalBench, RefCOCO, RoboCasa).
Training and evaluation configurations for every arm, including the \lora{} recipes and the robotics setup, are in Appendix~\ref{app:impl}; disclosure notes on seeds, selection, and missing cells are in \S\ref{subsec:setup}.
Code and trained prompt checkpoints will be released, including a minimal, runnable notebook (\texttt{code/soft2hard\_synthetic\_arithmetic.ipynb}) reproducing the synthetic-arithmetic soft-to-hard illustration of Appendix~\ref{app:synth_arith} end to end outside the full pipeline.

\bibliography{references_claude,references_mendeley,references_findings,references_extra}
\bibliographystyle{iclr2027_conference}

\newpage
\appendix

\section{Gradient Flow Comparison}
\label{app:gradflow}

\begin{figure}[t]
  \centering
  \includegraphics[width=\textwidth]{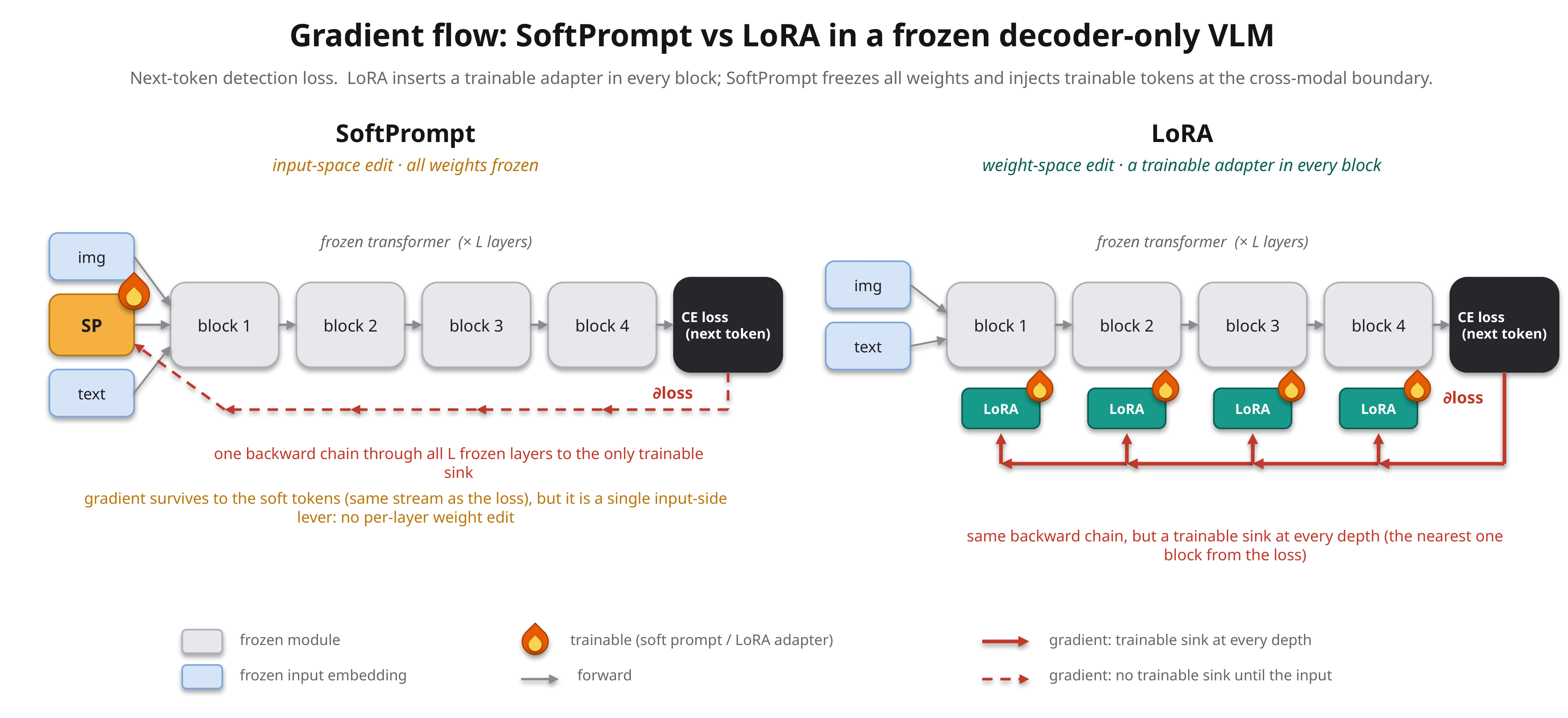}
  \caption{\textbf{A soft prompt is a weaker optimisation lever than \lora{}: the same backward chain, but a single trainable sink at its far end.}
    Gradient flow for the two adaptation methods on a frozen decoder-only VLM trained with the next-token detection loss.
    Gray boxes are \emph{frozen} modules and blue boxes frozen input embeddings; a flame marks a \emph{trainable} parameter (the amber soft-prompt tokens on the left, the teal \lora{} adapters on the right).
    Gray arrows are the forward pass; crimson arrows the backward chain from $\partial$loss, which traverses the same frozen blocks in both methods.
    \textbf{(Left, \ours{})} the chain crosses all $L$ frozen layers before reaching the only trainable sink (dashed: no update until the input), the soft tokens injected at the cross-modal boundary.
    \textbf{(Right, \lora{})} the same chain feeds a trainable adapter at every depth (solid taps); the nearest sink is one block from the loss, and updates land at every layer.
    \ours{} trades this optimisation disadvantage for ${\sim}10^3$--$10^4\times$ fewer trainable parameters and a backbone left fully intact.}
  \label{fig:gradflow}
\end{figure}

Figure~\ref{fig:gradflow} contrasts the backward-pass optimisation signature of \ours{} against \lora{} (\S\ref{subsec:method_training}): the same backward chain through every frozen block, but a single trainable sink at its far end rather than one at every depth.

\section{Implementation Details}
\label{app:impl}

\textbf{\ours{} (detection).}
All experiments use PyTorch~2.x with \texttt{bfloat16} on NVIDIA A6000 GPUs.
AdamW, lr $5{\times}10^{-3}$, cosine schedule, batch size 4, up to 5 epochs, space-token initialisation.
Measured over all 226 ablation runs, the training loop's single-GPU wall-clock in the 5-epoch regime (excluding vLLM export and test evaluation) has median $2.0$ hours (p10 $7$ minutes, p90 $5.5$ hours): the fastest domain (aerial-airport) trains in ${\sim}4$ minutes, while high-resolution domains (paper-parts) take up to $15$ hours.
Evaluation uses a vLLM engine with the trained soft-prompt rows appended to the embedding table as extended vocabulary.

\textbf{\lora{} SFT baseline (detection).}
Adapters on the language-model decoder's \texttt{q\_proj}, \texttt{k\_proj}, \texttt{v\_proj}, \texttt{o\_proj}, \texttt{gate\_proj}, \texttt{up\_proj}, \texttt{down\_proj} via \texttt{peft}; $\alpha{=}2r$, no dropout; lr $5{\times}10^{-5}$, cosine schedule, warmup ratio $0.2$, effective batch size 32, 30 epochs on 8 A6000s; best checkpoint by validation mAP; \texttt{image\_max\_pixels} $=1048576$.

\textbf{Robotics ($\pi_{0.5}$ / RoboCasa).}
Base policy: $\pi_{0.5}$ pretrained with RoboCasa data (PandaOmron embodiment; three camera views, 16-dim state, 12-dim action).
Tasks: \texttt{PickPlaceCounterToCabinet}, \texttt{TurnOnSinkFaucet}, \texttt{OpenDrawer}, each with its own $K{=}10$ per-seed demonstration subsets and normalisation statistics.
Soft prompt: learnable tokens on the PaliGemma prefix (prefix-only, $N{=}8$) or on both the prefix and the action-expert input (dual, $N{=}16$, $M{=}16$); IST format; the PaliGemma-side tokens use space-token initialisation as in \S\ref{subsec:method_init}, but the action-expert tokens are initialised randomly, since the action expert has no token vocabulary to seed from; trained with gradient accumulation to an effective batch of 128 and lr $2{\times}10^{-1}$ for the dual configuration, 2{,}000 steps.
The early-stopped \texttt{OpenDrawer} variant halts the same dual configuration at 750 steps; the stopping point was selected on a seed-0 sweep over steps and learning rate, then re-run on 3 seeds.
\lora{}: openpi reference configuration, rank 16 on attention and FFN modules; the ``+ action expert'' variant restricts to the rank-16 adapters (26.5M), and the PaliGemma-only variant matches the prefix prompt's coverage (19.6M).
Both arms train on the same $K{=}10$ demonstration episodes per seed.
Evaluation: 20 closed-loop episodes per seed on freshly sampled held-out scene instantiations; the environment's random generator is re-seeded per episode index, so all arms face byte-identical scenes.

\section{Per-Dataset Configurations}
\label{app:best_config}

Table~\ref{tab:best_config} lists, for every \rftwenty{} dataset, the frozen baseline, the selected \ours{} configuration behind Table~\ref{tab:rf20_categories}'s IST row (prompt length $L$, the epoch actually reached in training, and test mAP), and the best-of-position configuration (winning ordering, its $L$, and test mAP).
The selected lengths are $L \in \{1,2,3\}$ (mean $1.75$, \ie{} $7{,}168$ trainable parameters on average, range $4{,}096$--$12{,}288$); TSI wins 6 of 20 datasets, concentrated on single-class and text-dominant domains (\S\ref{subsec:position}).
All other hyperparameters are shared across datasets (Appendix~\ref{app:impl}).
The IST row uses dataset-specific budgets: 13 of 20 datasets train for 5 epochs (matching the \S\ref{sec:ablations} ablation setting), and 7 train for 50 epochs with early stopping (patience 15).

\begin{table}[h]
  \centering
  \caption{\textbf{IST wins 14 of 20 datasets; TSI takes the other 6, concentrated on single-class and text-dominant domains.}
    Per-dataset \ours{} configurations on \rftwenty{} (10-shot, mAP@50:95).
    IST: the fixed-position row of Table~\ref{tab:rf20_categories} (per-domain $L$).
    \textbf{Ep}: epoch actually reached by the IST checkpoint.
    Best of IST/TSI: the per-dataset best ordering (test-split selection; \S\ref{subsec:setup}, disclosure ii).}
  \label{tab:best_config}
  \small
  \setlength{\tabcolsep}{5pt}
  \begin{tabular}{llcc|cc|ccc}
    \toprule
    & & & & \multicolumn{2}{c|}{\textbf{IST}} & \multicolumn{3}{c}{\textbf{Best of IST/TSI}} \\
    \textbf{Dataset} & \textbf{Category} & \textbf{Baseline} & \textbf{Ep} & $L$ & mAP & Format & $L$ & mAP \\
    \midrule
    actions                   & Sports      &  2.9 &   5 & 3 &  3.2 & IST & 3 &  3.2 \\
    aerial-airport            & Aerial      &  4.4 &  41 & 1 &  9.1 & TSI & 8 & 12.1 \\
    all-elements              & Documents   & 10.6 &  15 & 1 & 13.4 & TSI & 1 & 14.4 \\
    aquarium-combined         & F.\,\&\,F.  & 29.2 &   7 & 1 & 35.7 & IST & 1 & 35.7 \\
    defect-detection          & Industrial  &  4.9 &   5 & 1 &  6.3 & TSI & 1 &  6.7 \\
    dentalai                  & Medical     &  0.0 &   3 & 2 &  0.1 & IST & 2 &  0.1 \\
    flir-camera-objects       & Other       &  9.3 &   5 & 2 & 16.6 & IST & 2 & 16.6 \\
    gwhd2021                  & F.\,\&\,F.  &  1.3 &  21 & 1 &  4.2 & TSI & 1 &  5.8 \\
    lacrosse-object-detection & Sports      & 14.2 &   5 & 1 & 24.3 & IST & 1 & 24.3 \\
    new-defects-in-wood       & Other       &  5.9 &   5 & 3 & 12.0 & IST & 3 & 12.0 \\
    orionproducts             & Other       &  3.1 &   5 & 3 &  6.0 & IST & 3 &  6.0 \\
    paper-parts               & Documents   &  4.4 &   5 & 2 & 15.9 & IST & 2 & 15.9 \\
    recode-waste              & Industrial  & 17.0 &   5 & 1 & 16.7 & IST & 1 & 16.7 \\
    soda-bottles              & Other       & 15.3 &  23 & 1 & 14.2 & IST & 1 & 14.2 \\
    the-dreidel-project       & Other       & 12.1 &   1 & 3 & 22.9 & IST & 3 & 22.9 \\
    trail-camera              & F.\,\&\,F.  & 41.9 &   5 & 2 & 41.9 & TSI & 2 & 43.5 \\
    water-meter               & Industrial  &  2.6 &   5 & 1 &  2.9 & IST & 1 &  2.9 \\
    wb-prova                  & F.\,\&\,F.  & 27.6 &   5 & 2 & 28.8 & IST & 2 & 28.8 \\
    wildfire-smoke            & Aerial      &  8.3 &  41 & 1 &  8.9 & TSI & 1 & 12.3 \\
    x-ray-id                  & Medical     &  0.1 &   5 & 3 &  0.0 & IST & 3 &  0.0 \\
    \midrule
    \textbf{Mean}             &             & 10.8 & & 1.75 & 14.2 & & & 14.7 \\
    \bottomrule
  \end{tabular}
\end{table}

\section{Gemma-4 Generalisation: Search Methodology and Full Results}
\label{app:gemma4}

\S\ref{subsec:gemma4} trains the method fresh on \texttt{Gemma-4-12B-it}, a VLM family unrelated to \qwen{}, and finds that while the recipe generalises (beats baseline on 5 of 8 above-floor pool datasets), the per-domain hyperparameter search used to pick each dataset's configuration does not transfer its own holdout signal to test as reliably as the equivalent \qwen{} search does.
This appendix gives the search methodology and the complete per-dataset accounting behind that finding.

\textbf{Search methodology.}
For each dataset, using only its 10-shot support set (never test), we hold out a fixed fraction as an internal validation split and run support-holdout successive halving over five stages: (S1) a format $\times$ length screen ($\{$IST, TSI$\}$ $\times$ $L \in \{1,2,3\}$, plus an $L{=}0$ control, space initialisation, lr $5{\times}10^{-3}$, no gradient accumulation); (S2) a gradient-accumulation sweep (accumulation over $4$ steps or the full support set) on the top-2 S1 configurations; (S3) a learning-rate refinement ($\{1{,}3{,}10\}{\times}10^{-3}$) of the best config so far; (S4) an initialisation check (\texttt{xavier}, \texttt{mean}) against space initialisation; and, added after the first pass on \qwen{} showed a wider net was needed for this backbone, (S5) an epoch-budget refinement ($10$, $15$ epochs against the default $5$) and an extended length screen up to $L{=}5$ for two datasets (defect-detection, flir-camera-objects) whose initial holdout margin over control was thin or, on a first attempt, entirely lost to transient GPU contention on a shared cluster node (confirmed by re-running in isolation).
Every stage uses the same holdout-mAP selection rule as the \qwen{} search (Appendix~\ref{app:impl}); the only difference from that search is the wider hyperparameter net (S1--S5 here vs.\ length alone there), made necessary because Gemma-4's holdout margins were consistently thinner at the default setting.
The selected config is then trained once more at full budget and evaluated on the true test split -- the number in Table~\ref{tab:gemma4} and below.

\begin{table}[!ht]
  \centering
  \caption{\textbf{The search's holdout margin does not predict its test margin.}
    Per-dataset search-selected configuration and the full holdout-vs-test accounting behind Table~\ref{tab:gemma4} and Table~\ref{tab:gemma4}'s finding box.
    \textbf{Test (untuned)}: an $L{\in}\{1,2,3\}$ IST, space-init, lr\,$5{\times}10^{-3}$, no-accumulation config with no search at all, for reference.
    Accum: gradient accumulation steps (\texttt{f} = full support set per step).
    All mAP@50:95, test-split numbers are best of two independent evaluations of the selected config (retrained from scratch; and the search-time checkpoint re-exported and re-evaluated), matching \S\ref{subsec:setup}'s disclosure that transfer-study checkpoints are test-selected for the ceiling but never for the headline claim.}
  \label{tab:gemma4_full}
  \small
  \setlength{\tabcolsep}{3.5pt}
  \resizebox{\textwidth}{!}{%
  \begin{tabular}{lccccc|cc|ccc}
    \toprule
    \textbf{Dataset} & $L$ & \textbf{Fmt} & \textbf{Init} & \textbf{LR} & \textbf{Accum}
                     & \textbf{Hold. ctrl} & \textbf{Hold. sel.}
                     & \textbf{Test base} & \textbf{Test sel.} & \textbf{Test (untuned)} \\
    \midrule
    actions                    & 1 & IST & space  & 5e-3 & 1 & 6.5  &  13.8 &  0.8 & \best{ 1.0} &  0.5 \\
    aerial-airport              & 1 & IST & mean   & 1e-3 & 1 & 2.9  &   3.4 & 11.3 & \best{11.4} &  9.8 \\
    all-elements               & 3 & TSI & space  & 5e-3 & 1 & 9.9  &  14.7 & 13.5 & \best{17.4} & 10.6 \\
    defect-detection           & 5 & TSI & space  & 1e-3 & 4 & 2.8  &  13.7 &  6.6 &  3.7        & \best{5.6} \\
    dentalai (floor)            & 1 & TSI & space  & 1e-3 & f & 0.9  &   9.9 &  0.5 &  0.6        &  0.4 \\
    flir-camera-objects        & 5 & TSI & space  & 5e-3 & 4 & 11.7 &  16.5 & 14.0 & \best{13.9} &  13.7 \\
    lacrosse-object-detection  & 2 & TSI & space  & 5e-3 & f & 9.0  &  18.5 & 11.3 &  10.5        & \best{13.1} \\
    new-defects-in-wood        & 2 & IST & xavier & 1e-3 & f & 8.9  &  13.2 &  3.7 &  3.9         & \best{4.2} \\
    wb-prova                   & 3 & TSI & space  & 1e-2 & 4 & 2.5  &  48.0 & 18.5 & \best{24.3}  & 19.0 \\
    \bottomrule
  \end{tabular}}
\end{table}

Two things follow from Table~\ref{tab:gemma4_full} that are not visible from the win/loss tally alone.
First, on defect-detection, lacrosse-object-detection, and new-defects-in-wood, the search-selected configuration is not merely a weak win, it is beaten on test by the untuned default we never selected on anything -- the search actively picks the wrong configuration for these three datasets, not just an insufficiently good one.
Second, the size of the holdout margin carries no information about test reliability: wb-prova has the largest holdout margin in the pool ($48.0$ vs.\ $2.5$ control, $19{\times}$) and the largest test gain ($+5.8$); defect-detection has the second-largest holdout margin ($13.7$ vs.\ $2.8$, $4.9{\times}$) and the worst test result in the pool.
A held-out mAP sweep over $10$ support images per class is evidently enough signal to separate configurations that generalise to the model's own training distribution, but not always enough to predict which configuration the frozen backbone will actually prefer on unseen test images -- a distinction the \qwen{} ablations (\S\ref{sec:ablations}) do not surface, because there the two happen to agree.
We report this as an open reliability gap in the search procedure itself, not as evidence against the underlying method: the same $10$-shot budget that lets \S\ref{sec:ablations}'s \qwen{} search reliably rank configurations does not, on this second backbone, carry enough signal to always rank them correctly.

\section{Serving Soft Prompts with Standard Inference Engines}
\label{app:vllm}

A practical objection to soft prompts is that they seem to require a custom forward pass (an embedding hook at the injection position), locking inference to a research codebase.
Our evaluation pipeline shows this is not the case: a trained soft prompt can be served by an \emph{unmodified} high-throughput engine, vLLM in our case, through the model's own vocabulary.

The export works as follows.
After training, the $L$ learnt vectors are appended as $L$ new rows of the backbone's input-embedding matrix, and $L$ new special tokens (\texttt{<SP\_0>}, \texttt{<SP\_1>}, \ldots) are registered in the tokenizer, one per row; the result is written as a standard checkpoint.
The base weights and the original embedding rows are untouched: the model gains $L$ rows and loses nothing.
At inference, the detection query simply contains the placeholder string \texttt{<SP\_0>}\ldots at the chosen injection position (\eg{} between the image and the text for IST), which tokenises to exactly the $L$ new ids (verified at load time); the engine embeds them like any other tokens, which \emph{is} the soft prompt.
No engine modification, no custom hooks, and all of the engine's optimisations (paged attention, tensor parallelism, batched generation) apply unchanged.

Three consequences are useful in practice.
First, one exported engine serves both arms of any soft-vs-base comparison: a query that omits the placeholder tokens is processed exactly as by the base model, which is how the distillation baseline of \S\ref{subsec:distill} shares a single engine with the soft-prompted arm.
Second, the marginal cost of a domain is $L$ embedding rows, so shipping an adapted domain is a vocabulary patch rather than a model copy: Table~\ref{tab:storage} measures this directly from the checkpoint files rather than from parameter counts alone, since on-disk size also reflects serialization overhead that a raw parameter count does not.
Third, switching domains is cheaper in kind, not just in size.
Injection is a lookup into the frozen embedding matrix, so activating a different prompt means addressing a different span of embedding rows, not attaching, merging, or hot-swapping a weight delta inside every target layer the way \lora{} serving must; our reference implementation (\texttt{code/softprompt\_pkg}) demonstrates the minimal case, detaching one trained prompt's hook and re-attaching another's in a single call, with no change to the frozen backbone's weights.
The same property extends to the exported checkpoint above: because each domain occupies a disjoint, fixed span of token ids, multiple domains' rows can coexist in one resident vocabulary, and a request activates whichever domain's tokens it includes.
All main-paper detection evaluations use this path.
Code, including the export and evaluation pipeline, will be released.

\begin{table}[h]
  \centering
  \caption{\textbf{The trained checkpoint is proportionally as small as the parameter count suggests, measured, not estimated.}
    On-disk size of the trainable adapter only (no base-model weights): \ours{}'s \texttt{torch.save} tensor vs.\ \lora{}'s \texttt{adapter\_model.safetensors}, both at their native training precision (fp32).
    \ours{} sizes are exact measurements at each $L$; the mean-$L{=}1.75$ row interpolates linearly (each additional token costs a fixed $16$\,KB, $=d{\times}4$ bytes for $d{=}4096$), matching the per-dataset $L$ mix used in Table~\ref{tab:rf20_categories}.
    Ratio: \lora{} size / \ours{} mean-$L$ size.
    \best{Bold}: $r{=}64$, the rank that matches \ours{}'s accuracy (Table~\ref{tab:rf20_categories}).}
  \label{tab:storage}
  \small
  \begin{tabular}{lrr}
    \toprule
    Checkpoint & Size & Ratio vs.\ \ours{} (mean $L$) \\
    \midrule
    \ours{} ($L{=}1$)              &  17.6\,KB & -- \\
    \ours{} ($L{=}2$)              &  33.6\,KB & -- \\
    \ours{} ($L{=}3$)              &  49.6\,KB & -- \\
    \ours{} (mean $L{=}1.75$)      &  29.6\,KB & $1\times$ \\
    \midrule
    \lora{} ($r{=}4$)              &  41.7\,MB & $1{,}442\times$ \\
    \lora{} ($r{=}8$)              &  83.3\,MB & $2{,}882\times$ \\
    \lora{} ($r{=}16$)             & 166.6\,MB & $5{,}762\times$ \\
    \lora{} ($r{=}32$)             & 333.1\,MB & $11{,}522\times$ \\
    \lora{} ($r{=}64$)             & \best{666.1\,MB} & \best{$23{,}041\times$} \\
    \lora{} ($r{=}128$)            &   1.3\,GB & $46{,}080\times$ \\
    \bottomrule
  \end{tabular}
\end{table}

The storage ratio at $r{=}64$ ($23{,}041\times$) is close to but slightly below the raw parameter-count ratio ($174.6\text{M}/7{,}168 \approx 24{,}357\times$, \S\ref{sec:method}): both checkpoints are fp32, so the gap is entirely serialization overhead, a fixed \texttt{torch.save} pickle header that is negligible against \lora{}'s multi-megabyte files but not against \ours{}'s few-KB ones.
We report the measured figure rather than the parameter estimate because it is what a deployment actually pays.

\section{Full \lora{} Rank Sweep}
\label{app:lora_sweep}

Table~\ref{tab:lora_full} reports both \lora{} variants at every rank.
The \emph{+ instructions} rows are the instruction-augmented protocol variant (\S\ref{subsec:main}); \ours{} has no counterpart in that setting yet (an instruction-augmented soft prompt is future work), so the main comparison (Table~\ref{tab:rf20_categories}, Figure~\ref{fig:pareto}) uses the default class-name protocol and the variant is shown here.
For reference, \ours{} with per-dataset choice of IST/TSI (position selected on the test split; \S\ref{subsec:setup}, disclosure ii) reaches $14.7$.

\textbf{\lora{} training configuration.} Every rank is trained per-dataset with LLaMA-Factory~\citep{zheng2024llamafactory} on \qwen{}-8B, target modules \texttt{q\_proj}, \texttt{k\_proj}, \texttt{v\_proj}, \texttt{o\_proj}, \texttt{gate\_proj}, \texttt{up\_proj}, \texttt{down\_proj} (attention and MLP), $\alpha{=}2r$ (e.g. $\alpha{=}128$ at $r{=}64$), dropout $0$, AdamW, learning rate $5\text{e-}5$, cosine schedule with $0.2$ warmup ratio, effective batch size $2$ (per-device batch $1$, gradient accumulation $2$), $70$ optimizer steps, bf16, seed $42$ — matching this paper's own default training seed (\S\ref{subsec:setup}). Verified directly against the saved checkpoints' own \texttt{adapter\_config.json} and \texttt{training\_args.bin} (not re-derived or estimated), confirmed identical across every sampled rank and dataset.

\begin{table}[h]
  \centering
  \caption{\textbf{\lora{} rank sweep on \rftwenty{} 10-shot (mAP@50:95): accuracy peaks at $r{=}32$--$64$ and declines at $r{=}128$.}
    Both variants (class names only; with per-class instructions).
    Figure~\ref{fig:pareto} plots the class-names sweep against its NaturalBench forgetting cost.
    \best{Bold}: best configuration per variant.}
  \label{tab:lora_full}
  \small
  \setlength{\tabcolsep}{4pt}
  \begin{tabular}{lcccccccc|c}
    \toprule
    \textbf{Method} & \textbf{Params} & \textbf{Aerial} & \textbf{Doc.}
                    & \textbf{F.\,\&\,F.} & \textbf{Ind.}
                    & \textbf{Med.} & \textbf{Sports}
                    & \textbf{Other} & \textbf{All} \\
    \midrule
    \lora{} ($r{=}4$)        &  10.9M &  8.1 &  8.1 & 27.0 &  8.8 &  0.0 & 11.9 & 10.5 & 12.1 \\
    \lora{} ($r{=}8$)        &  21.8M & 11.3 &  7.3 & 28.7 &  8.3 &  0.2 & 10.3 & 11.1 & 12.7 \\
    \lora{} ($r{=}16$)       &  43.6M & 12.3 & 13.8 & 28.1 &  6.6 &  0.3 &  8.6 & 10.9 & 12.8 \\
    \lora{} ($r{=}32$)       &  87.3M & 17.1 & 18.7 & 28.6 &  7.6 &  0.4 & 10.4 &  8.7 & 13.7 \\
    \lora{} ($r{=}64$)       & 174.6M & 20.2 & 19.9 & 28.6 &  7.1 &  0.7 &  9.6 &  9.4 & \best{14.2} \\
    \lora{} ($r{=}128$)      & 349.2M & 15.6 & 19.7 & 25.5 &  8.8 &  0.6 & 11.1 & 10.0 & 13.6 \\
    \midrule
    \lora{} + instr.\ ($r{=}4$)   &  10.9M &  8.5 &  9.4 & 26.9 & 10.8 &  0.3 & 12.2 & 12.1 & 13.1 \\
    \lora{} + instr.\ ($r{=}8$)   &  21.8M & 10.8 &  9.7 & 26.8 &  9.6 &  0.9 & 14.8 & 12.4 & 13.5 \\
    \lora{} + instr.\ ($r{=}16$)  &  43.6M & 12.7 & 13.0 & 29.4 & 11.3 &  0.3 & 10.3 & 10.6 & 13.9 \\
    \lora{} + instr.\ ($r{=}32$)  &  87.3M & 15.0 & 17.8 & 27.9 & 12.8 &  0.5 & 10.8 & 11.8 & \best{14.8} \\
    \lora{} + instr.\ ($r{=}64$)  & 174.6M & 15.6 & 20.8 & 26.7 &  9.2 &  1.0 & 10.3 & 12.1 & 14.5 \\
    \lora{} + instr.\ ($r{=}128$) & 349.2M & 14.7 & 19.0 & 27.7 &  7.4 &  0.7 & 12.4 & 10.9 & 14.0 \\
    \bottomrule
  \end{tabular}
\end{table}

\section{Single-Class Evaluation Baselines}
\label{app:single_class}

DetPO's strongest numbers use a single-class protocol: the VLM is queried once \emph{per class} per image, which multiplies inference cost by the class count and is not comparable to the single-call multi-class protocol used by every method in Table~\ref{tab:rf20_categories}.
For reference, DetPO reaches $15.3$ all-domain mAP under single-class evaluation and $17.5$ with VQA-score reranking.
VQA-score reranking is orthogonal to the prompt and applies equally to \ours{}; combining them is left for future work.

\begin{table}[h]
  \centering
  \caption{\textbf{DetPO's single-class protocol reaches $17.5$ mAP with VQA-score reranking, above \ours{}'s $14.2$ multi-class, at class-count$\times$ the inference cost.}
    Single-class evaluation (one class per VLM call; reference only, not cost-comparable to Table~\ref{tab:rf20_categories}).}
  \label{tab:single_class}
  \small
  \setlength{\tabcolsep}{4pt}
  \begin{tabular}{lcccccccc|c}
    \toprule
    \textbf{Method} & \textbf{Params} & \textbf{Aerial} & \textbf{Doc.}
                    & \textbf{F.\,\&\,F.} & \textbf{Ind.}
                    & \textbf{Med.} & \textbf{Sports}
                    & \textbf{Other} & \textbf{All} \\
    \midrule
    DetPO~\citep{Gare2026DetPO:Detection}     & 0 &  8.3 & 19.1 & 30.3 & 13.7 &  0.1 & 14.2 & 12.2 & 15.3 \\
    DetPO + VQA Score                         & 0 & 12.3 & 24.2 & 32.3 & 13.5 &  0.2 & 17.2 & 14.3 & 17.5 \\
    \bottomrule
  \end{tabular}
\end{table}

\newpage

\section{Cross-Task Grounding: Full RefCOCO Results}
\label{app:refcoco}

\begin{table}[!ht]
  \centering
  \caption{\textbf{Low-rank \lora{} does not measurably forget grounding; the advantage of \ours{} here is the guarantee, not the gap.}
    Cross-task forgetting on the RefCOCO family, acc@0.50 (\%).
    $\Delta$ is \lora{} (fine-tuned on the-dreidel-project, $r{=}4$) minus the frozen backbone.
    \ours{} is identical to the frozen backbone by construction (no prompt applied off-domain).}
  \label{tab:refcoco_forgetting}
  \small
  \setlength{\tabcolsep}{4pt}
  \resizebox{\textwidth}{!}{%
  \begin{tabular}{lcccccccc|c}
    \toprule
    Method
      & \makecell{RefCOCO\\val} & \makecell{RefCOCO\\testA} & \makecell{RefCOCO\\testB}
      & \makecell{RefCOCO+\\val} & \makecell{RefCOCO+\\testA} & \makecell{RefCOCO+\\testB}
      & \makecell{RefCOCOg\\val} & \makecell{RefCOCOg\\test} & Mean \\
    \midrule
    Frozen backbone               & 91.3 & 93.1 & 87.6 & 85.9 & 90.1 & 80.7 & 88.2 & 88.0 & 88.1 \\
    \lora{} (dreidel, $r{=}4$)    & 91.3 & 92.9 & 87.5 & 85.8 & 89.9 & 80.8 & 88.3 & 88.1 & 88.1 \\
    \midrule
    $\Delta$ (\lora{} $-$ base)   & $+0.00$ & $-0.10$ & $-0.10$ & $-0.06$ & $-0.16$ & $+0.10$ & $+0.10$ & $+0.18$ & $-0.01$ \\
    \ours{}                       & \multicolumn{9}{c}{$\equiv$ frozen backbone (identical by construction)} \\
    \bottomrule
  \end{tabular}}
\end{table}


\section{Cross-Model Transfer: Per-Category Results}
\label{app:transfer}

\begin{table}[!ht]
  \centering
  \caption{\textbf{Transfer helps most on Sports ($+5.3$) and Documents ($+3.3$), and hurts on Industrial ($-2.9$).}
    Cross-model transfer of trained soft prompts per \rftwenty{} super-category (\qwen{}-8B $\to$ Qwen3.5-9B, 10-shot, mAP@50:95).
    The 8B-trained prompt is loaded into Qwen3.5-9B with no retraining or projection.}
  \label{tab:transfer_full}
  \small
  \setlength{\tabcolsep}{5pt}
  \begin{tabular}{lccccccc|c}
    \toprule
    \textbf{Method} & \textbf{Aerial} & \textbf{Doc.} & \textbf{F.\,\&\,F.}
                    & \textbf{Ind.} & \textbf{Med.} & \textbf{Sports}
                    & \textbf{Other} & \textbf{All} \\
    \midrule
    Qwen3.5-9B baseline       & 8.8 & 7.4 & 26.0 & 4.3 & 0.1 & 4.5 & 8.1 & 9.9 \\
    9B + transferred prompt   & 9.0 & 10.7 & 25.6 & 1.3 & 0.1 & 9.8 & 9.9 & 10.7 \\
    8B origin (upper bound)   & 12.2 & 15.1 & 28.5 & 8.8 & 0.0 & 13.8 & 14.3 & 14.7 \\
    \midrule
    $\Delta$ vs.\ 9B baseline & $+0.2$ & $+3.3$ & $-0.4$ & $-2.9$ & $+0.1$ & $+5.3$ & $+1.8$ & $+0.8$ \\
    $\Delta$ vs.\ 8B origin   & $-3.2$ & $-4.4$ & $-2.8$ & $-7.5$ & $+0.1$ & $-4.0$ & $-4.5$ & $-4.0$ \\
    \bottomrule
  \end{tabular}
\end{table}

\section{Distillation vs. Direct Transfer: Per-Dataset Results}
\label{app:distill_transfer}

Table~\ref{tab:distill_transfer_full} gives the per-dataset numbers behind Table~\ref{tab:distill_transfer}: the same \qwen{}-8B-trained soft prompt, transferred to Qwen3.5-9B either as raw embeddings (direct, no retraining) or as the \S\ref{subsec:distill} hard prompt distilled from it (best of the global/per-class variant, no soft tokens at all).

\begin{table}[!ht]
  \centering
  \caption{\textbf{Direct embedding transfer beats its own distilled description on 18 of 20 datasets.}
    \qwen{}-8B $\to$ Qwen3.5-9B, test mAP@50:95, 10-shot. "Distilled" is the better of the \S\ref{subsec:distill} global/per-class hard prompt, evaluated with no soft tokens; "direct" loads the trained embeddings as-is (as in Appendix~\ref{app:transfer}). $\Delta$ is direct $-$ distilled.}
  \label{tab:distill_transfer_full}
  \small
  \setlength{\tabcolsep}{6pt}
  \begin{tabular}{lccc}
    \toprule
    \textbf{Dataset} & \textbf{Distilled} & \textbf{Direct} & $\Delta$ \\
    \midrule
    actions                    &  1.6 &  2.5 & $+1.0$ \\
    aerial-airport             &  7.8 & 10.7 & $+3.0$ \\
    all-elements                &  9.8 & 11.2 & $+1.4$ \\
    aquarium-combined          & 21.2 & 27.4 & $+6.2$ \\
    defect-detection           &  3.6 &  3.2 & $-0.4$ \\
    dentalai                   &  0.1 &  0.2 & $+0.1$ \\
    flir-camera-objects        & 14.7 & 17.6 & $+2.9$ \\
    gwhd2021                   &  1.8 &  0.9 & $-0.9$ \\
    lacrosse-object-detection  &  7.1 & 17.1 & $+9.9$ \\
    new-defects-in-wood        &  3.7 &  7.4 & $+3.8$ \\
    orionproducts               &  3.5 &  4.6 & $+1.1$ \\
    paper-parts                 &  6.1 & 10.2 & $+4.1$ \\
    recode-waste                &  7.5 &  9.8 & $+2.3$ \\
    soda-bottles                &  3.6 &  9.0 & $+5.4$ \\
    the-dreidel-project         & 15.9 & 17.0 & $+1.0$ \\
    trail-camera                & 46.3 & 46.4 & $+0.1$ \\
    water-meter                 &  0.4 &  0.7 & $+0.4$ \\
    wb-prova                    & 26.2 & 27.8 & $+1.6$ \\
    wildfire-smoke               &  5.5 &  7.6 & $+2.1$ \\
    x-ray-id                    &  0.0 &  0.1 & $+0.0$ \\
    \midrule
    \textbf{Mean}                &  9.3 & 11.6 & $+2.3$ \\
    \bottomrule
  \end{tabular}
\end{table}

The two datasets where distillation wins (defect-detection, gwhd2021) are both near-floor swaps below $4$ mAP either way; every dataset with a meaningful gap favours direct transfer, up to $+9.9$ on lacrosse-object-detection.

\section{Token-Substitution Probe: Discretising the Trained Prompt}
\label{app:substitution}

If the trained soft tokens have meaningful discrete neighbours (\S\ref{subsec:retrieval}), those neighbours should work as a prompt.
This probe tests that directly: at evaluation time, each trained soft token is replaced by its nearest neighbour (cosine, top-1) in the model's own representation spaces, either the text-vocabulary embeddings (\texttt{text}) or the cached frozen-encoder patch embeddings (\texttt{patch}), and the discretised prompt is evaluated in place of the trained one.
No retraining is involved; each dataset uses its canonical checkpoint.

\begin{table}[h]
  \centering
  \caption{\textbf{The trained prompt's discrete shadow carries most of its signal, more reliably in text than in image-patch space.}
    Substituting every trained token with its nearest text token keeps the prompt above the frozen baseline on 6 of 8 above-floor datasets and wins on every one of them; nearest-patch substitution is consistently weaker and collapses generation entirely (zero parseable detections) on three datasets.
    \best{Bold}: better substitution mode per dataset.
    dentalai is the floor control: neither the trained prompt nor its discrete shadow moves it.
    $^{\dagger}$wb-prova's nearest-patch substitution collapses generation entirely.}
  \label{tab:substitution}
  \small
  \setlength{\tabcolsep}{5pt}
  \begin{tabular}{llcccc}
    \toprule
    Dataset & Config & Baseline & Trained soft & Nearest \texttt{text} & Nearest \texttt{patch} \\
    \midrule
    aerial-airport            & 1T TSI &  4.4 & 11.4 & \best{6.9}  & 6.2 \\
    actions                   & 3T IST &  2.9 &  4.0 & \best{2.6}  & 0.0 \\
    all-elements              & 1T TSI & 10.6 & 14.4 & \best{11.1} & 11.0 \\
    defect-detection          & 1T IST &  4.9 &  8.0 & \best{5.8}  & 5.7 \\
    dentalai                  & 1T IST &  0.0 &  0.0 & 0.0         & 0.0 \\
    flir-camera-objects       & 2T IST &  9.3 & 16.6 & \best{11.3} & 0.0 \\
    lacrosse-object-detection & 2T IST & 14.2 & 24.0 & \best{17.6} & 17.0 \\
    new-defects-in-wood       & 2T IST &  5.9 & 11.0 & \best{8.7}  & 8.1 \\
    wb-prova$^{\dagger}$      & 2T IST & 27.6 & 28.8 & \best{25.1} & 0.0 \\
    \midrule
    \textbf{Mean}             &        &  8.9 & 13.1 & \best{9.9}  & 5.3 \\
    \bottomrule
  \end{tabular}
\end{table}

Two observations (Table~\ref{tab:substitution}).
First, text substitution stays above the frozen baseline on 6 of 8 above-floor pool datasets, all but actions and wb-prova$^{\dagger}$, which happen to be the two datasets where the trained prompt's own gain over baseline is smallest to begin with ($+1.1$ and $+1.2$ mAP, versus $+3.1$ to $+9.8$ elsewhere), leaving the least room for a discretised approximation to still clear the baseline.
Where it does clear the baseline, text substitution retains roughly $33\%$ of the trained prompt's gain in aggregate across those six datasets.
The learned directions therefore point at genuinely useful discrete tokens, the functional counterpart of the retrieval reading in \S\ref{subsec:retrieval} and of the distillation result in \S\ref{subsec:distill}, though the effect is a partial recovery rather than a free one.
Second, text substitution consistently outperforms patch substitution, winning on every above-floor dataset (pool means $9.9$ vs.\ $5.3$); on three datasets (actions, flir-camera-objects, wb-prova$^{\dagger}$), patch substitution collapses generation entirely to $0.0$ mAP, zero parseable detections, while text substitution on the same three datasets still produces valid, if below-baseline on two of them, output.
The useful discrete signal lives more reliably in the text-vocabulary space than in the visual-patch space.

\section{A Minimal Soft2Hard Illustration: Synthetic Arithmetic}
\label{app:synth_arith}

The distillation result in \S\ref{subsec:distill} and the substitution probe above both study soft-to-hard behaviour inside the full RF20 vision pipeline, where the soft tokens, the image, and a class-name query are all present together.
To isolate the phenomenon from that complexity, we built a minimal, non-visual companion: freeze a small causal LM (\texttt{Qwen3-VL-2B-Instruct}, used purely as a text model here) and train $L{=}10$ continuous soft tokens, in the same cross-modal-boundary-style placement (tokens immediately before the payload, \S\ref{subsec:method_position}), to solve a synthetic arithmetic task -- given a short list of integers with no instruction text, predict their sum or product -- then ask the trained model, in plain English, what the tokens mean.
Full code and a runnable notebook reproducing every number below are released with the paper (\texttt{code/soft2hard\_synthetic\_arithmetic.py} / \texttt{.ipynb}).

\begin{table}[h]
  \centering
  \caption{\textbf{The soft prompt learns the synthetic task reliably across all four conditions.}
    Exact-match accuracy, 100 held-out examples, greedy decoding, 20 epochs, $L{=}10$.}
  \label{tab:synth_arith}
  \small
  \setlength{\tabcolsep}{8pt}
  \begin{tabular}{lc}
    \toprule
    Task & Exact match \\
    \midrule
    Addition, 2-digit operands (0--99)       & \best{1.00} \\
    Addition, 1-digit operands (0--9)        & 0.96 \\
    Multiplication, 2-digit operands (0--99) & 0.95 \\
    Multiplication, 1-digit operands (0--9)  & 0.86 \\
    \bottomrule
  \end{tabular}
\end{table}

Asked about the soft tokens in isolation (\texttt{"Elaborate: `<SP>'."}, no numbers), the model produces incoherent output in every condition: fragments of Hebrew, Greek, or other scripts unrelated to arithmetic.
A nearest-vocabulary-token check, the same probe as Appendix~\ref{app:substitution} applied here, explains why: across all 4 conditions ($40$ soft-token positions checked in total), the single nearest vocabulary neighbour by cosine similarity is, without exception, the same rare musical-notation glyph ($\text{cosine sim.} \approx 0.80$--$0.82$), regardless of task.
This is not operation-specific signal, just where a lightly-trained $10$-token prompt happens to sit in embedding space; the isolated query's incoherence is consistent with the model echoing something near that generic neighbour rather than reasoning about the span at all.

Asking about the tokens \emph{in the context they were trained in} -- immediately followed by a real number list, the exact shape seen during training -- changes this completely.
\texttt{"What is the task prompt of: `<SP>: 53, 93'?"} answers \emph{"What is the sum of the two numbers: 53 and 93?" \textbf{Answer:} $53+93=\mathbf{146}$} (correct, on the 2-digit addition prompt), and \emph{"Multiply 6 by 0." ... $6\times0=0$. \textbf{Answer: 0}} (correct, 1-digit multiplication).
\texttt{"Rewrite in your own words: ..."} reliably names the operation (\emph{"Multiply 53 by 93."}) without always attempting to solve it; \texttt{"Describe the task prompt of: ..."} is good but the least stable of the in-context templates, occasionally hallucinating an unrelated frame entirely.
Templates that invite ad-hoc computation rather than description (\texttt{"Elaborate"}, bare \texttt{"Rewrite"}) more often get the arithmetic itself wrong on 2-digit multiplication ($53\times93$ answered $4949$ or $4989$, not $4929$) even while still correctly implying the operation -- the exact trained template reaches $0.86$--$1.00$ exact match (Table~\ref{tab:synth_arith}), but placing the same tokens inside different surrounding phrasing does not inherit that reliability.
One further asymmetry: self-description accuracy does not track task-solving accuracy -- the 1-digit addition prompt solves the task at $0.96$ exact match but is the one condition where \texttt{"What is the task prompt of"} most often misreads it as a sequence-completion task rather than addition.

\begin{finding} \textbf{Finding.} In a minimal setting stripped of vision entirely, a trained soft prompt is not self-describing in isolation -- an isolated query only recovers the nearest generic vocabulary token, not task content -- but it \emph{is} soft-to-hard distillable once the query preserves the context it was trained in, at which point natural language recovers a real, often exactly correct, description of the task the tokens encode. This mirrors, in miniature, the RF20 result that verbalisation requires the tokens to be queried inside a working prompt, not in a vacuum.
\end{finding}

\section{Multi-Seed Variance Check}
\label{app:seed_variance}

Every ablation-table number in the main paper is a single run; \S\ref{subsec:setup}'s disclosure (i) estimates a $20$--$28\%$ seed-to-seed spread from this check. Each pool dataset's canonical Space/IST checkpoint (the same configuration reported in \S\ref{sec:ablations}) is retrained from two additional seeds (123, 2024) alongside the original (seed 42), with no other change.

\begin{table}[h]
  \centering
  \caption{\textbf{Seed-to-seed spread averages $28\%$ of the per-dataset mean, matching the $20$--$28\%$ estimate the main paper's ablation tables are built on.}
    Test mAP@50:95 across three training seeds per pool dataset, canonical Space/IST configuration.
    CV: coefficient of variation (std / mean), the seed-noise scale disclosure (i) reports.
    dentalai is the floor control (excluded from the summary, as in every other ablation table); 
    }
  \label{tab:seed_variance}
  \small
  \setlength{\tabcolsep}{7pt}
  \begin{tabular}{lccccc}
    \toprule
    Dataset & seed 42 & seed 123 & seed 2024 & Mean $\pm$ Std & CV \\
    \midrule
    aerial-airport                     &  7.98 &  6.45 &  8.10 & $7.51 \pm 0.92$  & $12\%$ \\
    actions                            &  3.94 &  5.40 &  3.93 & $4.42 \pm 0.84$  & $19\%$ \\
    all-elements                       & 11.59 &  6.81 & 14.97 & $11.12 \pm 4.10$ & $37\%$ \\
    defect-detection                   &  7.97 &  5.52 &  5.45 & $6.31 \pm 1.43$  & $23\%$ \\
    flir-camera-objects                & 16.55 & 16.96 &  7.00 & $13.50 \pm 5.64$ & $42\%$ \\
    lacrosse-object-detection          & 24.10 & 11.02 & 13.24 & $16.12 \pm 7.00$ & $43\%$ \\
    new-defects-in-wood                & 10.97 & 12.37 &  9.42 & $10.92 \pm 1.47$ & $14\%$ \\
    \midrule
    dentalai (floor control)           &  0.00 &  0.00 &  0.01 & \multicolumn{2}{c}{excluded} \\
    wb-prova$^{\dagger}$               & 28.80 & 13.74 & 19.64 & $20.73 \pm 7.59$ & $37\%$ \\
    \bottomrule
  \end{tabular}
\end{table}

The mean coefficient of variation across the eight fully-collected, above-floor datasets is $28\%$ (range $12$--$43\%$), confirming the $20$--$28\%$ disclosed in \S\ref{subsec:setup}: single-run ablation numbers carry real seed noise, and per-dataset deltas below ${\sim}1$ mAP should be read as ties rather than genuine effects.
The spread is not uniform: aerial-airport and new-defects-in-wood are comparatively stable ($12$--$14\%$), while all-elements, flir-camera-objects, lacrosse-object-detection, and wb-prova swing by $37$--$43\%$ of their own mean.
This does not track test-set size: flir-camera-objects has the pool's largest test split (500 images) and one of the highest CVs, while aerial-airport has the smallest (34 images) and the lowest, so whatever drives a given dataset's seed sensitivity is not simply "fewer test images."

\subsection{The spread is training noise, not evaluation noise}
\label{app:eval_seed_variance}

Table~\ref{tab:seed_variance} retrains from three seeds, so its spread conflates two sources: different random initializations reaching different local optima during training, and stochasticity in the vLLM generation step at test time. To isolate them, we take each pool dataset's seed-42 checkpoint (the one already trained above) and re-run test-set evaluation twice more, varying only the generation seed (\texttt{--skip\_training}, no retraining).

\begin{table}[h]
  \centering
  \caption{\textbf{Eval-time variance is near zero; the seed-to-seed spread in Table~\ref{tab:seed_variance} comes from training, not evaluation.} Same checkpoint, three evaluation seeds, test mAP@50:95. CV here is $\leq\!5\%$ on every dataset, an order of magnitude below the $12$--$43\%$ training-seed CVs.}
  \label{tab:eval_seed_variance}
  \small
  \setlength{\tabcolsep}{7pt}
  \begin{tabular}{lcccc}
    \toprule
    Dataset & eval-seed A & eval-seed B & eval-seed C & Std \\
    \midrule
    aerial-airport             & 7.98  & 7.39  & 7.36  & 0.35 \\
    actions                    & 3.94  & 3.94  & 3.96  & 0.01 \\
    all-elements                & 11.59 & 11.76 & 11.86 & 0.13 \\
    defect-detection            & 7.97  & 7.97  & 7.97  & 0.00 \\
    flir-camera-objects         & 16.55 & 16.13 & 16.56 & 0.24 \\
    lacrosse-object-detection   & 24.10 & 23.99 & 23.99 & 0.06 \\
    new-defects-in-wood         & 10.97 & 11.00 & 11.00 & 0.02 \\
    wb-prova$^{\dagger}$        & 12.05 & 12.93 & 12.95 & 0.51 \\
    \midrule
    dentalai (floor control)    & 0.00  & 0.00  & 0.00  & \multicolumn{1}{c}{excluded} \\
    \bottomrule
  \end{tabular}
\end{table}
$^{\dagger}$wb-prova's checkpoint is re-evaluated at \texttt{max\_pixels}$=3$M (matching Table~\ref{tab:seed_variance}'s canonical setting) for all three eval seeds; its own \texttt{init\_ablation} run used a different setting, so it is not the source of the value quoted elsewhere in the paper.

\begin{finding} \textbf{Finding.} Re-evaluating a fixed checkpoint under three generation seeds moves test mAP by at most $0.51$ points on any pool dataset (mean std $0.17$), against $12$--$43\%$ training-seed CVs on the same datasets (Table~\ref{tab:seed_variance}). The seed-to-seed spread this paper discloses is a training-time phenomenon: different soft-prompt initializations land in different local optima, not measurement noise at test time.
\end{finding}

\subsection{Is the spread specific to SoftPrompt?}
\label{app:lora_seed_variance}

The 10-shot regime itself could simply be noisy, in which case any adapter would show comparable run-to-run spread. To test this, we retrain the accuracy-matching \lora{} configuration ($r{=}64$, \S\ref{subsec:main}) from the same three seeds, same nine pool datasets, identical data and hyperparameters (Appendix~\ref{app:lora_sweep}) apart from \texttt{seed}, and evaluate test mAP with the same protocol used to produce every \lora{} number in this paper.

\begin{table}[h]
  \centering
  \caption{\textbf{\ours{}'s training-seed spread is not a general property of 10-shot \rftwenty{} training: it averages $7.9\times$ larger than \lora{}'s at the same accuracy-matching rank, on 6 of 8 above-floor pool datasets.} Test mAP@50:95, three training seeds ($42$, $123$, $2024$), $r{=}64$.}
  \label{tab:lora_seed_variance}
  \small
  \setlength{\tabcolsep}{6pt}
  \begin{tabular}{lccc|cc}
    \toprule
    & \multicolumn{2}{c}{Std (mAP pts)} & & & \\
    Dataset & \ours{} & \lora{} & Ratio & \ours{} mean & \lora{} mean \\
    \midrule
    aerial-airport             & 0.92 & 1.84 & 0.5 &  7.51 & 11.19 \\
    actions$^{\ddagger}$       & 0.84 & 0.40 & 2.1 &  4.42 &  8.08 \\
    all-elements                & 4.10 & 0.37 & 11.0 & 11.12 & 13.44 \\
    defect-detection            & 1.43 & 0.48 & 3.0 &  6.31 & 10.12 \\
    flir-camera-objects         & 5.64 & 0.15 & 36.9 & 13.50 &  4.99 \\
    lacrosse-object-detection   & 7.00 & 1.42 & 4.9 & 16.12 & 13.56 \\
    new-defects-in-wood         & 1.47 & 2.12 & 0.7 & 10.92 &  9.51 \\
    wb-prova                    & 7.59 & 1.80 & 4.2 & 20.73 & 27.75 \\
    \midrule
    dentalai (floor control)    & 0.01 & 0.04 & \multicolumn{1}{c}{--} & 0.00 & 0.38 \\
    \bottomrule
  \end{tabular}
\end{table}
$^{\ddagger}$actions' seed-42 \lora{} checkpoint (\texttt{qwen3vl-8b-rank64/actions/multi\_class}) is corrupted at the source (a $13.7$MB \texttt{adapter\_model.safetensors}, versus $\sim\!698$MB for every other dataset and seed; \texttt{safetensors} fails to deserialize it), so only seeds $123$ and $2024$ are available for this row.

\begin{finding} \textbf{Finding.} \lora{} at the accuracy-matching rank is markedly more stable across training seeds than \ours{}: on 6 of 8 above-floor pool datasets its std is $2.1$--$36.9\times$ smaller, averaging $7.9\times$ across all 8 (excluding dentalai's floor effect on both), and its eval-only variance is likewise negligible (Table~\ref{tab:eval_seed_variance}), ruling out measurement noise as the explanation. Two datasets (aerial-airport, new-defects-in-wood) are exceptions where \lora{}'s spread instead exceeds \ours{}'s. The dominant pattern nonetheless says the seed-to-seed spread disclosed throughout this paper is largely an artifact of \ours{}'s tiny parameter count, not a property of 10-shot \rftwenty{} training in general: a few thousand trainable prompt parameters land in more different local optima across random seeds than \lora{}'s $174.6$M do.
\end{finding}

\newpage

\section{Qualitative Detections}
\label{app:qual}

\begin{figure}[h]
  \centering
  \includegraphics[width=0.9\textwidth]{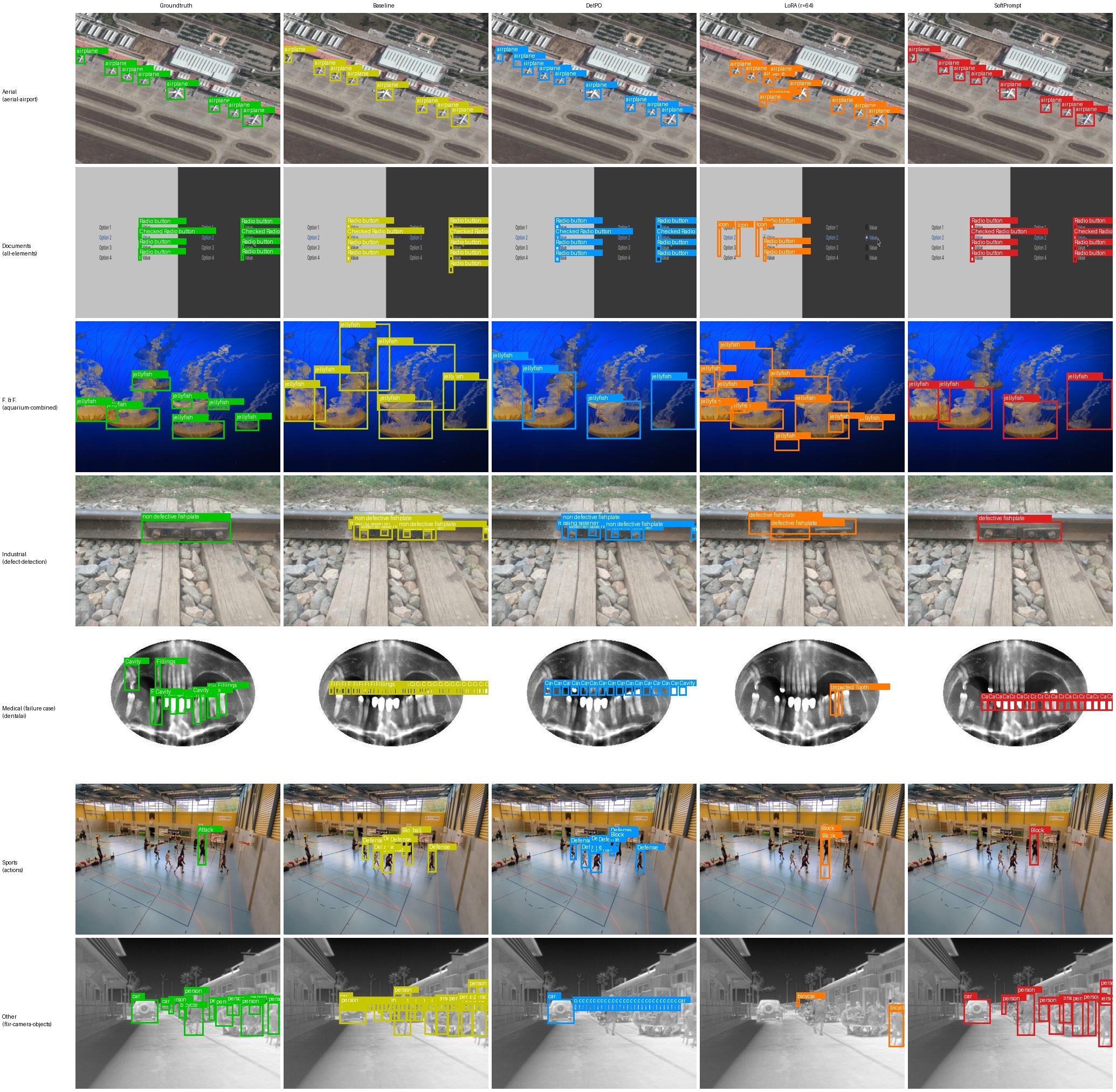}
  \caption{Qualitative detections on one held-out test image per super-category (7 of \rftwenty{}'s domains), comparing the frozen zero-shot baseline, DetPO, LoRA (rank 64), and SoftPrompt against groundtruth. The frozen baseline is the weakest column throughout, often over-generating or repeating a single label (dentalai, defect-detection); SoftPrompt localizes and labels cleanly on most domains but inherits the baseline's failure on dentalai, where no adapter recovers domain knowledge the frozen backbone never had.}
  \label{fig:qual}
\end{figure}

\end{document}